\documentclass[energies,article,accept,pdftex,moreauthors]{Definitions/mdpi} 
\firstpage{1} 
\pubvolume{1}
\issuenum{1}
\articlenumber{0}
\pubyear{2023}
\copyrightyear{2023}
\externaleditor{Antonio Gabaldón, María Carmen Ruiz-Abellón   and  Luis Alfredo Fernández-Jiménez}  
\datereceived{} 
 \daterevised{~} 
\dateaccepted{} 
\datepublished{} 
\hreflink{https://doi.org/} 

\usepackage[labelformat=simple]{subcaption}

\DeclareCaptionLabelFormat{subcaptionlabel}{\normalfont(\textbf{#2}\normalfont)}
\Title{Interval Load Forecasting for Individual Households in the Presence of Electric Vehicle Charging}

\TitleCitation{Interval Load Forecasting for Individual Households in the Presence of Electric Vehicle Charging}

\Author{Raiden 
 Skala $^{1}$, Mohamed Ahmed T. A. Elgalhud $^{1}$\orcidA{}, Katarina Grolinger $^{1,}$*\orcidB{} and Syed Mir $^{2}$} 

\AuthorNames{Raiden Skala, Mohamed Elgalhud, Katarina Grolinger, and Syed Mir}

\AuthorCitation{Skala, R.; Elgalhud, M.; Grolinger, K.; Mir, S.}

\address{%
$^{1}$ \quad  Department of Electrical and Computer Engineering, Western University, London, ON N6A 5B9, Canada; rskala@uwo.ca (R.S.); melgalhu@uwo.ca (M.A.T.A.E.)              \\
$^{2}$ \quad London Hydro, London, ON  N6A 4H6, Canada; mirs@londonhydro.com
}

\corres{Correspondence: kgroling@uwo.ca; Tel.: +1-519-661-2111 (ext. 81407)}

\abstract{
The transition to Electric Vehicles (EV) in place of traditional internal combustion engines is increasing societal demand for electricity. The ability to integrate the additional demand from EV charging into forecasting electricity demand is critical for maintaining the reliability of electricity generation and distribution. Load forecasting studies typically exclude households with home EV charging, focusing on offices, schools, and public charging stations. Moreover, they provide point forecasts which do not offer information about prediction uncertainty. Consequently, this paper proposes the Long Short-Term Memory Bayesian Neural Networks (LSTM-BNNs) for household load forecasting in presence of EV charging. The approach takes advantage of the LSTM model to capture the time dependencies and uses the dropout layer with Bayesian inference to generate prediction intervals. Results show that the proposed LSTM-BNNs achieve accuracy similar to point forecasts with the advantage of prediction intervals. Moreover, the impact of lockdowns related to the COVID-19 pandemic on the load forecasting model is examined, and the analysis shows that there is no major change in the model performance as, for the considered households, the randomness of the EV charging outweighs the change due to pandemic.}


\begin{document}


\section{Introduction}

Affordable and reliable sources of electricity enable the sustainable growth of strong economies and can improve the average person’s quality of life \cite{1su7054783} by providing reliable access to appliances, medical equipment, communication, entertainment, and other devices. The dependence on power grids to provide electricity is increasing due to the continuous integration of novel electronic devices into every aspect of modern life \cite{2energy.gov} as these devices rely on a reliable source of electricity. The mainstream adoption of electric vehicles (EVs) away from traditional internal combustion engine (ICE) vehicles for consumer use is set further to entrench reliance on access to electricity due to the increased electricity demand for charging. While this transition can be a positive step in reducing carbon emissions \mbox{\cite{3energy.gov}}, embracing EVs will shift transportation energy requirements from petroleum-based products to electric grids. Of specific interest to this paper, is the demand created by charging EVs in residential households, which is frequently utilized due to its relative affordability and convenience.

As countries, such as Canada, plan to ban the sale of new ICE vehicles by 2035 \cite{4ghunem_2022}, preparations are required to ensure the success of this shift. This includes actions such as installing charging stations, increasing electricity generation capacity, investing into battery technologies, and improving infrastructure throughout the grid to handle the higher loads required by EV charging. The capability of electricity distribution companies to accurately forecast the hourly electricity consumption of residential households that own EVs is instrumental in the transition to EVs as it assists the utility companies to anticipate and manage increased energy demand, plan sufficient capacity to meet expected demand, and ensure grid stability. Failure in predictive ability poses a risk to the balance between electricity supply and demand, which can impose serious threats to grid stability, human life, and overall economic interest  \mbox{\cite{5yamashita2008analysis}}. A loss in the balance between supply and demand can lead to power outages, brownouts, and other disruptions. In turn, these electricity disruptions can severely disrupt critical infrastructures including transportation, communications, and financial systems, as well as essential services such as emergency response and healthcare.

There have been extensive efforts to create predictive load forecasting models using machine learning (ML) with historical energy consumption data collected by smart meters or similar technologies, often combined with meteorological information 
\mbox{\cite{24s21217115, sehovac2020deep}}. In recent years, deep learning techniques, especially those based on Recurrent Neural Networks (RNNs) have been outperforming other techniques \mbox{\cite{23sun2022individualized, jung2021attention}}. While these studies had great successes in terms of accuracy for a variety of use cases, they do not specifically address short-term forecasting for individual households in presence of EV charging \cite{6al2019review} which introduces challenges due to variations of charging patterns. 

Moreover, most predictive models for load forecasting generate point predictions instead of an interval for their expected electricity demand \cite{6al2019review}, which limits the usefulness of the forecast in decision-making. By providing only a single value for expected electricity demand, these models fail to convey the range of potential outcomes and the degree of uncertainty associated with each prediction. Furthermore, providing only a point forecast, without the range values, may not offer sufficient information for effective risk management. In contrast, interval forecasting approaches provide decision-makers with a more nuanced understanding of the possible outcomes, enabling them to make more informed and effective decisions. For example, considering the full range of possible outcomes, instead of a single value, allows the stakeholders to plan for different scenarios and better mitigate~risks.  

To address these drawbacks, this paper proposes a probabilistic interval forecasting approach for predicting the hourly electricity demand in households with EV charging. By using probabilistic methods, our approach generates a range of likely outcomes rather than a single-point estimate which provides a more comprehensive understanding of the potential effects of EV charging on household electricity demand, gives information about the uncertainty associated with the predicted value due to dynamic charging behaviors, and offers decision-makers a more complete picture of the forecasted demand. The interval predictions are generated with Long Short-Term Memory Bayesian Neural Networks (LSTM-BNNs). LSTM was chosen as it is well-suited for capturing temporal dependencies in data while BNN was added to estimate the probability distribution of expected values for interval predictions. LSTM-BNN was trained using historical household electricity consumption data and local temperature data. To assess the effectiveness of the proposed LSTM-BNN model, its performance, measured using four metrics, is compared to the performance of the standard point prediction LSTM model. Additionally, due to the impact of the COVID-19 pandemic on electricity consumption patterns, the point and interval models have been examined on two datasets: one with the lockdown period and one without. The results show that the accuracy greatly varies among households, but for each household, the proposed LSTM-BNN achieves similar accuracy to point forecasts while providing the advantage of prediction intervals.

The remainder of the paper is organized as follows: Section 2 provides background information on LSTM and BNN techniques and introduces the four common performance measurements used for gauging the effectiveness of regression models while Section 3 reviews related work on load forecasting and interval forecasting. The proposed LSTM-BNN interval forecasting approach is described in Section 4 followed by the evaluation presented in Section 5. Finally, Section 6 concludes the paper.

\section{Background}

This section begins by introducing Long Short-Term Memory (LSTM) networks and Bayesian Neural Networks (BNNs), followed by a discussion of the four performance measures commonly used for assessing regression models.

\subsection{Long Short-Term Memory Neural Network}

Neural networks are a type of machine learning model inspired by the human brain: they use interconnected artificial neurons to learn and process information by mimicking the way biological neurons signal to one another \cite{yu2019review}. A recurrent neural network (RNN) is a type of neural network designed to process sequential data by using internal memory and recurrent connections, allowing it to capture temporal dependencies and patterns in the data. A Long Short-Term Memory (LSTM) neural network model is similar to RNN models in that it can capture temporal relationships by using an internal memory mechanism to keep track of past inputs and selectively remember or forget certain information. The main difference between an LSTM and RNN model is that LSTM models have additional structures, such as gating mechanisms, that provide better control over the flow of gradients and help prevent the vanishing and exploding gradient problems that can occur in standard RNNs, making them more effective for modeling longer sequences of data. LSTM computation at time $t$ is given as follows:

    \begin{align}
     f_{t} &= \sigma(W_{fx}x_{t} + W_{fh}h_{t-1} + b_{f})\\
     i_{t} &= \sigma(W_{ix}x_{t} + W_{ih}h_{t-1} + b_{i})\\
     o_{t} &= \sigma(W_{ox}x_{t} + W_{oh}h_{t-1} + b_{o})\\
     \bar C_{t} &=\varphi(W_{cx}x_{t} + W_{ch}h_{t-1} + b_{c})\\
      C_{t} &=f_{t}\odot C_{t-1} + i_{t}\odot \bar C_{t}\\
      h_{t} &= o_{t}\odot \varphi(C_{t})
      \end{align}\label{eq:lstm}

Equations (1)--(3) depict the computation at the forget $f_t$, input $i_t$, and output $o_t$ gates respectively, while Equations (4)--(6) determine the cell state $C_t$ and hidden state $h_t$. The sigmoid ($\sigma$) and tanh ($\varphi$) functions contribute to controlling exploding gradients by keeping values between zero to one and negative one to one respectively. The current cell input $x_t$ and previous cell hidden state $h_{(t-1)}$ are the inputs received by the LSTM cell. The gate biases $b_f$, $b_i$, $b_o$, and $b_c$, the current cell weight matrices $W_{fx}$, $W_{ix}$, $W_{ox}$, and $W_{cx}$ and the hidden state cell weight matrices $W_{fh}$, $W_{ih}$, $W_{oh}$, $W_{ch}$ of each LSTM cell are adjusted throughout the training process using backpropagation through time with the goal of minimizing the loss between the predicted and true values. The use of $\odot$ indicates computing the elementwise Hadamard product of two matrices. 

Due to its ability to capture temporal dependencies over long periods of time, the LSTM model has been very successful in many domains including load forecasting \mbox{\cite{8fekri2022distributed, jagait2021load}}. For the same reason, we use the LSTM cells in the proposed LSTM-BNN interval forecasting approach. 

\subsection{Bayesian Neural Network}

The Bayesian Neural Network (BNN) model \cite{25hastie2009elements} relies on Bayesian inference to determine the posterior predictive distribution with the ultimate goal of quantifying the uncertainty introduced by the models so as to explain the trustworthiness of the prediction. This is achieved by incorporating previous inputs $X$ and outputs $Y$ as well as model parameters $\omega$ in Bayes' theorem as follows:

\begin{equation}
    P(\omega|X, Y) =  \frac {P(Y|X, \omega)\cdot P(\omega)} {P(Y|X)} \label{eq:bay1}
\end{equation}

\noindent Here, $P()$ indicates the probabilities and $P(\cdot|\cdot)$ are conditional probabilities. 

By computing the integral of the full posterior distribution, given in (\ref{eq:bay1}), multiple times using different samples from the model parameters, a distribution can be generated for a predicted value $y_{new}$ using new inputs $x_{new}$:

\begin{equation}
    P(y_{new}|x_{new}, X, Y) = \int P(y_{new}|x_{new}, \omega)\cdot P(\omega|X,Y)d\omega    \label{eq:bay2}
\end{equation}

\noindent However, due to the full posterior probability being computationally demanding for deep neural networks, alternative approaches are required to make the use of Bayesian inference feasible in practice. Zhang and Mahadevan \cite{10zhang2020bayesian} demonstrated Monte Carlo dropout remaining active while a network generates predictions to be sufficient for approximating the posterior predictive distribution as it minimizes the relative entropy between the approximate and true posterior distributions while remaining computationally feasible. Consequently, our approach takes advantage of BNN and the dropout technique to generate interval load forecasts for households with EVs. 

\subsection{Performance Metrics} \label {sec:PerformanceMetrics}

The four prominent performance metrics that are used for evaluating the margin of error between a prediction made by a machine learning model and the true value are: Mean Absolute Percent Error (MAPE), Mean Square Error (MSE), Root Mean Square Error (RMSE), and Mean Absolute Error (MAE) \mbox{\cite{8fekri2022distributed,10zhang2020bayesian}}.While there are other metrics specifically dealing with evaluating probabilistic forecasts \mbox{\cite{wan2013probabilistic}}, we primarily use the mentioned metrics as they allow us to compare point and interval forecasts. These metrics are calculated according to the following equations:

\begin{align}
\text{MAPE} &=  \frac{100\% }{N} \sum_{i=1}^{N} \frac{\left| {y_{i} - \hat{y_{i}}} \right|}{y_{i}} \\ \label {eq:MAPE}
\text{MSE} &=  \frac{1}{N}\sum_{i=1}^{N} (y_{i} - \hat{y_{i}} )^2 \\ \label {eq:MSE}
\text{RMSE} &= \sqrt{(\frac{1}{N})\sum_{i=1}^{N}(y_{i} - \hat{y_{i}})^{2}}\\ 
\text{MAE} &=  \frac{1}{N}\sum_{i=1}^{N}\left | y_{i} - \hat{y_{i}} \right| 
\end{align}

\noindent where $y_i$ is the true value of the $i$-th sample, $\hat{y_{i}}$ is the predicted value for the $i$-th sample and $N$ is the total number of samples. 

MAPE has an advantage over the other three metrics as it is a scale-independent metric representing the error as a percentage of the actual value and therefore suitable for comparing models on datasets of different value scales. MSE and RMSE metrics are both based on the Euclidean distance to determine the level of error between predicted and true values. The difference between the MSE and RMSE metrics is that MSE provides more severe punishment for predictions that are very different from the true value. MAE is used to measure the mean absolute difference between predictions and true values and is less severe at penalizing large differences between predicted and true values than MSE or RMSE. To obtain a different view of forecasting accuracy, our study employs all four metrics.

\section{Related Work}

This section first reviews recent load forecasting studies focusing on those based on machine learning and then discusses techniques for interval predictions in different domains.

\subsection{Electricity Load Forecasting}

This subsection first reviews recent load forecasting studies for a diversity of consumers including residential households and buildings. This provides insights into state-of-the-art models 
 and represents directions for forecasting in the presence of EV changing. Next, related work in predicting EV charging in various settings is examined.

An LSTM-based model for short-term load forecasting on the individual household level was proposed by Kong et al. \cite{9kong2017short}. They \cite{9kong2017short} found that a significant hurdle to creating forecasts at the household level is the large degree of diversity and volatility in energy consumption between households when compared to making forecasts at the substation level. This difficulty in residential forecasting due to load variability and concept drift also aligns with the findings of Fekri et al. \cite{8fekri2022distributed, 12fekri2021deep}.

Residential load forecasting was also investigated by Zhang et al. \cite{13zhang2018forecasting}: while Kong~et~al. \cite{9kong2017short} used LSTM-based approach  Zhang~et~al. \cite{13zhang2018forecasting} employed Support Vector Regression (SVR). In their study, Zhang~et~al. \cite{13zhang2018forecasting} investigated predicting daily and hourly electricity consumption for 15 households with data obtained from smart meters. The accuracy of the load predictions varied significantly across households, depending on the variability of energy-related behaviors among occupants. Daily load estimates were generally more accurate, as they mitigated the randomness in hourly changes.

L’Heureux et al. \cite{22l2022transformer} presented tranformer-based architecture for electrical load forecasting. They adapted the transformer model from the Natural Language Processing (NLP) domain for load forecasting by modifying the NLP transformer workflow, adding N-space transformation, and designing a novel technique for handling contextual features. They examined the proposed transformer-based architecture on 19 different data streams and with four different forecasting horizons. For most data streams and forecasting horizons, the transformer accuracy was better than the Seq2Seq network; however, for 12-h forecast Seq2Seq was slightly better

A multi-node load forecasting was investigated by Tan~et~al. \mbox{\cite{tan2022multi}}: they proposed multi-task learning combined with a multi-modal feature module based on an inception-gated temporal convolutional network for node load prediction. The feature extraction module captures the coupling information from the historical data of the node, while the multi-task learning utilizes a soft sharing mechanism to leverage the shared information across nodes to improve the forecast accuracy. Experimental results demonstrate the effectiveness of the proposed method in accurately forecasting load demand across multiple nodes.

Ribeiro~et~al. \mbox{\cite{ribeiro2022short}} investigated short- and very short-term load forecasting for warehouses and compared several machine learning and deep learning models including linear regression, decision trees, artificial neural networks, and LSTM models. In their experiments RNN, LSTM, and GRU cells achieved comparable results.
Jian~et~al. \mbox{\cite{jiang2022very}} also worked on very short-term load forecasting: they proposed a framework based on an autoformer which combines decomposition transformers with auto-correlation mechanism. Multi-layer perceptron layers are added to the autoformer for an improved deep information extraction. In their experiments, the proposed deep-autoformer framework outperformed several deep-learning techniques on the task of very short-term residential load forecasting.

An encoder-decoder RNN architecture with a dual attention mechanism was proposed by Ozcan~et~al. \cite{24s21217115} to improve the performance of the RNN model. The attention mechanism in the encoder helps identify important features whereas the attention in the decoder assists the context vector and provides longer memory. In their experiments, the encoder-decoder RNN architecture achieved improved accuracy in comparison to LSTM; however, the computation complexity was increased.   

Short-term load forecasting has been investigated by Sun~et~al. \cite{23sun2022individualized}: they proposed a framework based on LSTM and an enhanced sine cosine algorithm (SCA). The authors enhanced the performance of the SCA, a meta-heuristic method for optimization problems, by incorporating a chaos operator and multilevel modulation factors. In experiments, they compared the modified SCA with several other population intelligence algorithms including particle swarm optimization and the whale optimization algorithm and showed that SCA improves performance. 

There are very few studies concerned with load forecasting for EV charging demand and they mostly consider scenarios such as parking lots, fleets, and regional demand. For example, Amini~et~al. \cite{11amini2016arima} investigated forecasting of EV charging demand for parking lots. Their approach used an Autoregressive Integrated Moving Average (ARIMA) model with driving patterns and distances driven as inputs to determine the day-ahead demand of the conventional electrical load and charging demand of EV parking lots. Two simulated test systems, 6-bus and IEEE 24-bus systems, were used to examine the effectiveness of the proposed approach.

Yi~et~al. \mbox{\cite{yi2022electric}} highlighted the importance of accurate demand forecasting for planning and management of electric vehicle charging infrastructure. They presented a deep learning-based method for forecasting the charging demand of commercial EV charging stations by utilizing LSTM as a base for the Seq2Seq model and combining it with a clustering technique. The evaluation on over 1200 charging sites from the State of Utah and the City of Los Angeles showed that the proposed method outperforms other forecasting models such as ARIMA, Prophet, and XGBoost.
For forecasting EV charging demand at charging stations in Colorado, Koohfar~et~al. \mbox{\cite{koohfar2023prediction}} proposed a transformer-based deep learning approach. The proposed approach was compared to time-series and machine learning models including ARIMA, SARIMA, LSTM, and RNN. While for longer time horizons the transformer outperformed other techniques, for short-term forecasting (7 days ahead), LSTM and transformer achieved comparable results.

 A multi-feature data fusion technique combined with LSTM was proposed by Aduama~et~al. \mbox{\cite{aduama2023multi}} to improve the EV charging station load forecasting. They generate three sets of inputs for LSTM consisting of load and weather data pertaining to different historical periods. These three sets of data are then passed to the LSTM models which generate three predictions, and, finally, the LSTM outputs are combined using a data fusion technique. In their experiments, the proposed fusion-based approach achieved better accuracy than traditional LSTM in predicting EV charging station demand.

Zheng~et~al. \cite{14zheng2020systematic} were interested in predicting the overall load from EVs in the city of Shenzhen, China. They recognize the diversity of charging patterns and therefore break down the fleet into four groups: private EVs, taxis, busses, and official EVs. Their approach provides a mid-and-long term EV load charging model based on the current utilization of EVs in Shenzhen using probabilistic models for EV charging profiles and forecasting EV market growth in the city using the Bass model. As they are concerned with the regional EV demand, some of the randomnesses of the individual EV charging is remedied through aggregation. Similarly, Arias and Bae \cite{15ARIAS2016327} considered forecasting load for groups of EVs. Specifically, they take advantage of historical traffic data and
weather data to formulate the forecasting model. First, traffic patterns are classified, then factors influencing traffic patterns are identified, and finally, a decision tree formulates the forecasting model.

Strategies for handling growing EV charging demand were investigated by \linebreak \mbox{Al-Ogaili~et~al. \cite{6al2019review}}. They classify the EV control strategies into scheduling, clustering, and forecasting strategies recognizing that precise estimates of charging are critical for fault prevention and network stability. They note that the stochastic nature of EV charging demand requires advanced forecasting techniques, commonly combined with the need for extensive data including historical charging data, weather, and travel patterns, which may not be readily available. Forecasting studies Al-Ogaili~et~al. \cite{6al2019review} examined include predictions for groups of EVs or geographical regions, charging stations, and specific types of EVs (e.g., busses).

The reviewed studies \cite{9kong2017short, 8fekri2022distributed, 12fekri2021deep, 13zhang2018forecasting, 22l2022transformer, 24s21217115, 23sun2022individualized,tan2022multi,jiang2022very,ribeiro2022short} on generic load forecasting represent the state-of-the-art in energy forecasting but their behavior in presence of EV charging has not been examined. Nevertheless, they represent a great foundation for forecasting EV charging load. On the other hand, EV-related studies \cite{11amini2016arima, 14zheng2020systematic, 15ARIAS2016327, 6al2019review,yi2022electric,koohfar2023prediction,aduama2023multi} do consider EV charging but they do so for a group of EVs, parking lots, charging-station, or regions, and do not confider forecasting load for individual households in presence of EVs. In contrast, we focus on predicting power consumption for individual households in presence of EV charging. Moreover, in contrast to point predictions provided in the aforementioned studies, our study offers interval predictions.    

\subsection{Interval Predictions}

This subsection reviews approaches that have been taken by authors across different domains to create regression models that provide an interval for predictions. In contrast to point predictions, interval predictions quantify uncertainties and provide additional information for decision-making.    

Interval predictions were generated for electricity spot pricing by Maciejowska~et~al. \cite{16maciejowska2016probabilistic} for the British power market using factor quantile regression averaging. First, point predictions are obtained with a collection of models including autoregressive models, threshold autoregressive models,  semiparametric autoregressive models, neural networks, and others. Next, point predictions generated by the mentioned models are combined using quantile regression averaging to provide final interval forecasts. The proposed approach performed better than the benchmark autoregressive model.

Shi~et~al. \cite{17shi2017direct} considered interval predictions for forecasting wind power generation to quantify uncertainties in renewable energy generation. They train an RNN model with two outputs, one for the upper and one for the lower bound of a regression interval of predictions using the Lower and Upper Bound Estimation (LUBE) method. A new cost function incorporating prediction interval was designed and the dragonfly algorithm was introduced to tune the parameters of the RNN prediction model. One of the major challenges associated with training neural networks using the LUBE method is the difficulty in achieving convergence and occasionally the model may not converge \cite{18KABIR2021106878}. Consequently, Kabir~et~al. \cite{18KABIR2021106878} developed a customizable cost function to improve the convergence of LUBE models and assist in constructing prediction intervals with neural networks.

Zhang and Mahadevan \cite{10zhang2020bayesian} proposed interval forecasting for flight trajectory prediction and safety assessment by combining deep learning with uncertainty characterized by a Bayesian approach. Two types of Bayesian networks (BNN), feedforward neural networks and LSTM networks, are trained from different perspectives and then blended to create final predictions. In both BNNs, the dropout strategy quantifies model prediction uncertainty. The BNN approach was also successful in the work of Niu and Liang \cite{19niu2018nuclear} where they improve nuclear mass and single-neutron separation energy prediction accuracy for determining nuclear effective reactions. In their experiments, Niu and Liang \cite{19niu2018nuclear} demonstrate that a Bayesian approach can be combined with various forecasting techniques to improve nuclear mass predictions.

The reviewed studies \cite{16maciejowska2016probabilistic,17shi2017direct,18KABIR2021106878, 10zhang2020bayesian, 19niu2018nuclear} created interval predictions with various machine learning and statistical methods in various domains; however, none of them considered forecasting household electricity load in the presence of EV charging.  
Like our study, the works of Zhang and Mahadevan \cite{10zhang2020bayesian} and  Niu and Liang \cite{19niu2018nuclear} also employed BNN techniques to create interval prediction but they used it for very different use cases than load forecasting (flight trajectory \cite{10zhang2020bayesian} and nuclear mass predictions \cite{19niu2018nuclear}).


\section{Interval Load Forecasting in Presence of EV Charging}
This section presents the problem formulation and methodology of the proposed interval forecasting for household load prediction in the presence of EV charging. The approach uses only historical energy consumption data obtained from smart meters and weather data which makes it practical and scalable for real-world applications as there is no need to collect data regarding EV charging habits or EV specifications.

\noindent{\textbf{Problem Statement: 
} 
Consider a time series of historical data for a household with EV charging represented as a sequence of input-output pairs $(x_t, y_t)$ where $x_t$ is a vector of features describing the state of the electricity consumption at time $t$ including contextual factors such as temperature, time of day, day of the week, and day of the year, and $y_t$ is a vector of real-valued electricity consumption values for this household at time $t$. The goal is to learn a probabilistic model $p(\hat{y}_{t+1}|x_{t+1},D)$, where $D$ represents historical observations and $\hat{y}_{t+1}$ represents the predicted value, that can predict the output for a new input with uncertainty quantification represented as an interval $I$.  

This interval is created by generating multiple predictions through different network configurations to obtain the Bayesian approximation of the predicted value (interval center) as shown in Equation \mbox{\eqref{eq:ExpectedPredict}}. The  minimum and the maximum of the interval are computed as shown in Equations \mbox{\eqref{eq:Imin}} and \mbox{\eqref{eq:Imax}} respectively: 

\begin{equation}
E[\hat{y}_{t+1}]=\frac{1}{N}\sum_{i=1}^N\hat{y}_{t+1}^i
\label{eq:ExpectedPredict}
\end{equation}
\begin{equation}
I_{min}=E[\hat{y}_{t+1}]-\sigma_{\hat{y}_{{t+1}}} 
\label{eq:Imin}
\end{equation}
\begin{equation}
I_{max}=E[\hat{y}_{t+1}]+\sigma_{\hat{y}_{{t+1}}}
\label{eq:Imax}
\end{equation}

\noindent where $N$ is the number of predictions generated for the time step $(t+1)$ and $\sigma_{\hat{y}_{{t+1}}}$ is the standard deviation of the predictive distribution computed as follows:

\begin{equation}
\sigma_{\hat{y}_{{t+1}}}=\sqrt{\frac{\sum_{i=1}^N(\hat{y}_{t+1}^i-E[\hat{y}_{t+1}])^2}{N}}
\label{eq:std}
\end{equation}

The overall interval forecasting process is shown in Figure \mbox{\ref{fig:methodology}}, while details of each component are described in the following subsections.

\begin{figure}[H]
\begin{adjustwidth}{-\extralength}{0cm}
\centering
\includegraphics[width=17.5cm]{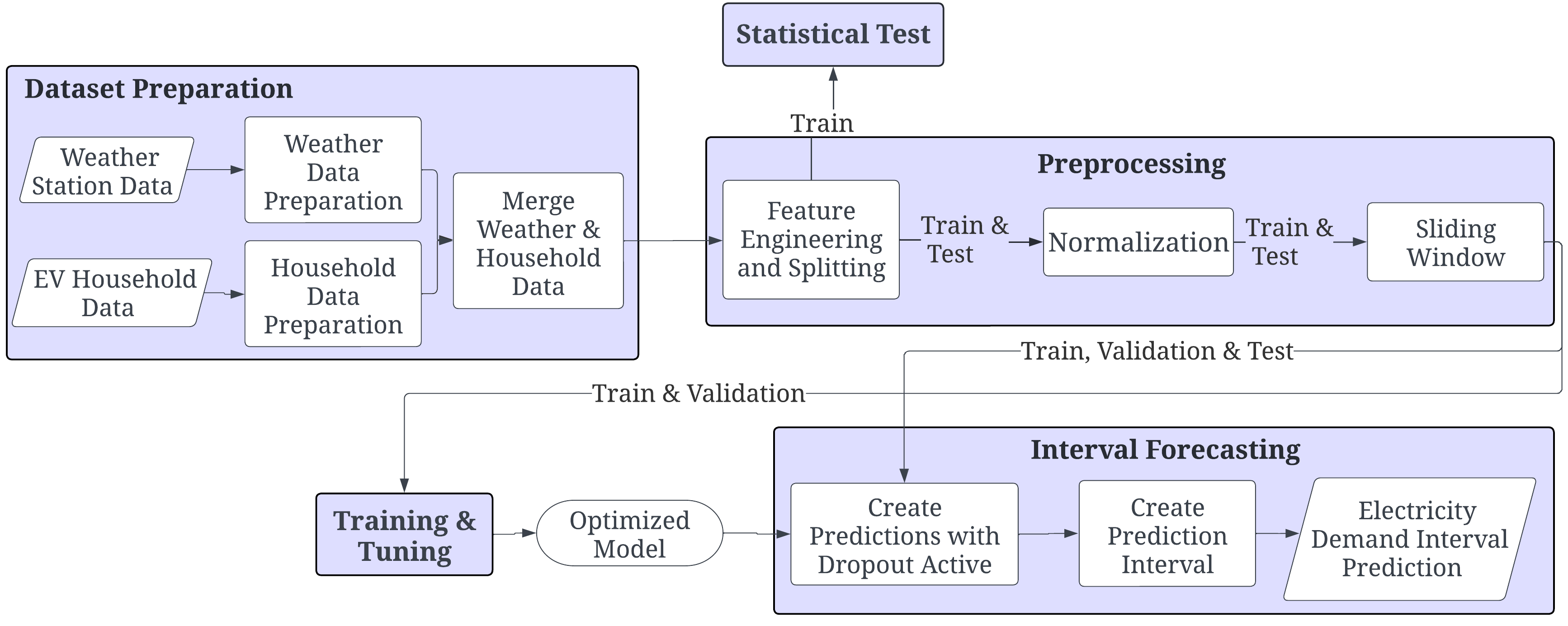}
\end{adjustwidth}

\caption{Overview of the proposed LSTM-BNNs interval forecasting approach for household load prediction in presence of EV charging. \label{fig:methodology}}
\end{figure}

\subsection{Dataset Preparation} \label{sec:DatasetPreparation}

The two types of datasets being used are weather station data and historical household electricity consumption data. Each dataset undergoes preparation individually before they are merged.

The weather station data consists of multiple datasets from multiple weather stations in the approximate geographical areas surrounding the EV household. The features used from the weather station data are the hourly timestamps and the temperature recordings as the temperature is often considered the most influential weather factor in load forecasting \cite{mirasgedis2006models}. The Weather Data Preparation conducted on the individual weather station datasets and shown in Figure \ref{fig:methodology} includes filling in missing temperature readings and combining all the weather station data. Missing temperature readings are filled using weighted averaging of the nearest complete temperatures. Since beyond the city, there are no additional geographical details given for any of the EV households, the temperature data from all stations are combined by averaging the temperatures from several weather stations to create a single average temperature dataset. In order to match the timestamps in the EV Household datasets, the timestamps in the average temperature dataset are adjusted to adhere to daylight savings time (DST).

The household data here refers to hourly data recorded by the smart meter or a similar device. The two initial features in this data are the consumption period and the electricity consumed within that period. The consumption period for the household data initially contains both the start date and time, and the end date and time of the current electricity consumption period. These data, as indicated in Figure \ref{fig:methodology}, undergo Household Data Preparation which involves isolating and only keeping the start date-time of the consumption period so that it can be merged with the weather station data. The electricity consumption feature from the initial dataset remains unchanged.

As part of the processing for both the weather station and household data, an additional time feature is generated. This feature is necessary because at the end of DST each year, the time is set backward one hour, resulting in two instances of the same date-time. This creates a conflict in merging the weather station and household data using only the date-time feature, as there are duplicate non-unique date-times that have no distinguishing differences. The additional time feature is added to both the household and weather station data to indicate if the specific date-time falls in DST or not. This removes the merging ambiguity for the duplicate date-times, as only the first occurrence of the date-time will occur during DST.

After the initial preparations are completed, each of the individual household datasets is merged with the average temperature dataset using the date-time and the additional time feature. In preparation for machine learning, the merged dataset proceeds to the preprocessing step. 

\subsection{Preprocessing} \label{sec:preprocessing}

After the weather and household datasets are merged, the datataset undergoes the following preprocessing steps: feature engineering, splitting the data into train and test sets, and normalizing the train and test sets. Feature engineering step takes advantage of the recorded interval start date-time from the original household data to generate nine features as shown in Table \ref{tab:features}. The purpose of creating additional features is to provide context information to the machine learning model, enabling it to generate better predictions.     

\begin{table}[H] 
\caption{Features extracted from the reading date and~time.\label{tab:features}}
\setlength{\cellWidtha}{\textwidth/2-2\tabcolsep-0.5in}																	
\setlength{\cellWidthb}{\textwidth/2-2\tabcolsep+0.5in}	 																
\scalebox{1}[1]{\begin{tabularx}{\textwidth}{>{\raggedright\arraybackslash}m{\cellWidtha}>{\raggedright\arraybackslash}m{\cellWidthb}}																	

\toprule
\textbf{Date-Time Feature}	& \textbf{Feature Description}\\
\midrule 
Day of week	& Range from 0 to 6 starting on Monday\\
Quarter of year    & Range from 1 to 4 starting with January to March			\\
Month of year    & Range from 1 to 12 starting in January			\\
Day of the Year   & Range from 1 to 365 (366 for leap year)			\\
Day of the Month   & Range from 1 to 31 \\
Week of the Year   & Range from 0 to 53 for weeks starting on Monday\\
Hour of the Day   & Range from 0 to 23 starting at 12:00AM\\
Year   & Range from 2018 to 2020 \\
DST   & Value of 1 (not DST) or 0 (DST)	\\

\bottomrule \label{tab:features}
\end{tabularx}}
\end{table}
\vspace{-6pt}
%
%

Following feature engineering, the dataset was split into training and testing sets: the last 10\% of readings are assigned to the test set and the remaining to the training set. 
Then, a portion of the training set was separated to use as the validation set for model selection. As a result of the validation set creation, the distribution of the data becomes 80\% for training, 10\% for validation, and ~10\% for testing.

Next z-score normalization was used to reduce the dominance of large features and improve convergence. This technique was chosen over other normalization techniques as it is good at handling outliers present during peak electricity consumption events. The z-score normalization transforms the feature to the mean of 0 and the standard deviation of 1 as follows:

\begin{equation}
    z_{ij}= \frac{x_{ij}-\mu_j} {\sigma_j} \\ 
    \label{eq:ZScore}
\end{equation}

\noindent where $x_{ij}$ and $z_{ij}$ are the initial unscaled and scaled values of the $i$-th sample of the $j$-th feature respectively, and $\mu$ and $\sigma$ are the mean and standard deviation of all the samples of the $j$-th feature respectively. Note that the mean and standard deviations are calculated only on the training set to avoid data leakage. 

Next, the sliding window technique is employed to prepare data for the machine learning model and to provide the model with a fixed number of previous electricity consumption, time-date, and temperature features as the inputs for predicting the next time step. This is accomplished by creating an input window that contains all features including the time-step, temperature, and consumption data within the window size \cite{12fekri2021deep}. For instance, for a window of size $w$, electricity consumption together with all other features for the past $w$ time steps are used as the input for predicting the next energy consumption values. The window slides for $s$ steps to create the next sample. The advantage of the windowing technique for electricity forecasting is in allowing the model to consider the demand for recent time steps when making predictions. The exact window size $w$ is determined within the optimization process. 

The siding window technique is applied to each training, validation, and test set. After this step, the samples have the dimension of $w \times f$ where $f$ is the number of features.

\subsection{Training and Tuning} \label{sec:TrainingTuning}

As shown in Figure \ref{fig:methodology}, the Training and Tuning stage follows the Preprocessing step. The deep learning technique LSTM was selected as the machine learning model because in recent years it has demonstrated great successes in load forecasting and outperformed other forecasting techniques \cite{12fekri2021deep, 8fekri2022distributed}. The hyperparameter search was carried out with Bayesian optimization as unlike grid or random search, this method performs a more directed exploration of a defined tuning space by selecting hyperparameters that lead to a local, ideally global, minimum loss \cite{20pmlr-v80-falkner18a}. The Bayesian optimization achieves a minimum loss by using the posterior distribution of the Mean Squared Error (MSE) loss function determined by previous models to guide the selection of new hyperparameter combinations. This directed selection process minimizes the time and computational resources needed for the exploration of the defined hyperparameter space \cite{20pmlr-v80-falkner18a}.

In this work, the search space explored included window size, batch size, the number of LSTM layers, the number of neurons in each LSTM layer, the learning rate for the Adam optimizer, and the dropout probability. The window size determines the number of previous time steps to be used as the input to the network while the batch size specifies the number of training windows a single batch contains. The number of LSTM layers and the number of LSTM neurons are adjusted to find a balance between increasing model complexity and variance to fit the training set while maximizing the model’s ability to generalize and make accurate predictions when given novel data points. The learning rate for the Adam optimizer is a critical parameter for training each model as it determines the rate at which updates are made to the weight and bias parameters of the model. Using a learning rate that is suitable for finding minima in the loss function enables the model to converge efficiently. 

Finally, the dropout probability hyperparameter is used to prevent overfitting: it determines the probability that neurons in a layer will randomly be given zero values. Using a dropout probability that is too high can have a detrimental effect on overall performance as it could result in too many inactive neurons and prevent the model from learning. As the dropout technique is typically used only to reduce overfitting to training data, it is typically disabled when the model is making predictions (inference time). However, within the proposed LSTM-BNN model, dropout is also active while making predictions as it is the key component to creating the probabilistic interval predictions as described in the following subsection. 

The LSTM model is trained and tuned using Bayesian optimization for each individual household independently. Once training and tuning are completed, the model is ready to proceed to the Interval Forecasting step.

\subsection{Interval Forecasting} \label{sec:IntervalForecasting}

This subsection described how the trained and tuned LSTM model is used to create the prediction intervals. The approach is inspired by the works of Zhang and Mahadevan \cite{10zhang2020bayesian} and  Niu and Liang \cite{19niu2018nuclear}, and as those works, it employs the BNN technique to generate intervals. However, they used the BNN technique for different applications and with different networks.  

With the active dropout, the trained model makes a sufficiently large number of predictions for each sample of the dataset. Due to dropout being active, the model has a high likelihood to produce a different point prediction each time it makes a prediction even though all inputs are the same. This variation in predictions is because while the dropout is active there is a probability that any component, excluding input and output neurons, can be removed from the prediction calculation. The varying point predictions due to the use of different components in prediction calculations allow for the construction of an interval prediction as a variational approximation of Bayesian inference for the model uncertainty \cite{10zhang2020bayesian}.

After multiple predictions were made for the same input samples, the mean and standard deviation for each sample is determined using the collection of point predictions the model created. Finally, the interval prediction is given as one standard deviation above and below the mean value of the point predictions for each sample.

A summary of the four steps taken in creating the interval prediction is given in the following steps:

\begin{enumerate} [label=\textbf{Step {\arabic*}:}]  
\item	Make multiple predictions for a given input.
\item	Compute the mean and standard deviation of the predictions for each input sample.
\item	Center the interval at the mean value.
\item   Define the upper bound of the interval as one standard deviation above the mean, and the lower bound as one standard deviation below the mean.
\end{enumerate}

\noindent All the models generated by Bayesian optimization are evaluated on the training and validation sets while only the best-performing model for each household selected on the validation set is evaluated on the test set. In other words, the model selection is carried out on the validation set. 

\subsection{Statistical Tests}

Household energy consumption is dependent on the behaviors of its occupant, and as such changes when those behaviors change. We are interested to examine the effect the COVID-19 related pandemic lockdowns had on households with EV charging. As the \mbox{Shapiro–Wilk} test showed that the datasets do not follow a normal distribution, \mbox{Mann–Whitney U test was used to examine }electricity consumption with and without lockdown in order to determine if lockdowns created a change in household electricity consumption habits to the extent that it is statistically different and could impact the predictive capacity of the model.

To carry out this analysis, two datasets are considered: the first dataset, referred here as the \textit{lockdown dataset
}, contains the entire original EV household dataset including data from the lockdown as well as before lockdowns. The second dataset, \textit{ non-lockdown dataset} is a subset of the EV household dataset containing only data collected outside of the lockdown period. Both the lockdown and non-lockdown datasets go through the same preparation, preprocessing, and prediction processes for the creation of the prediction model.

As seen in Figure \ref{fig:methodology}, evaluation using \mbox{Mann-Whitney} is carried out before the normalization is applied. The test is performed after the dataset is split into components training, validation, and testing. A comparison of the differences in the \mbox{Mann-Whitney} results helps us determine if there is a greater similarity between training and test sets} for the lockdown dataset or for the non-lockdown dataset which is a subset of the lockdown dataset.

There are three possible outcomes for the comparison of the \mbox{Mann-Whitney} results for the lockdown and non-lockdown datasets. The first is that there is no significant difference between training and test sets for either the lockdown or non-lockdown datasets. The second possible outcome is that there is a greater difference between the lockdown dataset training and test sets than for the non-lockdown dataset. And the third possibility is that there is a greater difference between the non-lockdown dataset training and test sets than for the lockdown dataset.

The \mbox{Mann-Whitney} results are compared with the final performance of the models that are trained on the lockdown and non-lockdown datasets to observe whether there is a correlation between differences in the datasets and model predictive performance. The analysis of the \mbox{Mann-Whitney} results and the predictive performance of the model will improve our understanding of the conditions under which the generated models are reliable. Understanding when a model is reliable is critical for mitigating the risks of a blackout because it ensures that decisions are made based on reliable forecasting information.  
\section{Evaluation}

This research was carried out in collaboration with London Hydro, a local electrical distribution utility for the city of London, Ontario, Canada. The real-world dataset provided by London Hydro was shared through Green Button Connect My Data (CDM), a platform for secured sharing of energy data with the consumer’s consent. Through work like this, London Hydro is preparing for the increased proliferation of EVs and the corresponding increase in electricity demand. London Hydro needs home EV charging data to identify nonwire solutions such as scheduling charging during off-peak hours 
 when there is solar~generation.

In this evaluation, we consider four households with EVs and refer to them as EV1, EV2, EV3, and EV4. The time period ranges for all four households' recordings are very similar, as given in Table \ref{tab:datasetDates}. Note that the time period ranges non-lockdown dataset are the same for all four households. The Weather Station Data was obtained from Environment and Climate Change Canada and consists of two datasets from two weather observation stations in the London area that were merged by averaging their temperature readings, as discussed in Section \ref{sec:preprocessing}.

For comparison of lockdown to non-lockdown data, additional four subsets of the EV household electricity consumption datasets are created by removing the lockdown data following the start of lockdowns on 1 March 2020. For each of the four households, 
 two individual LSTM-BNN predictor models were trained and tuned. The first model for each household is trained and evaluated on the entire dataset which contains lockdown and non-lockdown electricity consumption data, and the second model for each household is trained and evaluated using only the non-lockdown electricity consumption data.

\begin{table}[H] 
\caption{Data ranges for lockdown and non-lockdown~datasets.\label{tab:datasetDates}}
\setlength{\cellWidtha}{\textwidth/3-2\tabcolsep-0.8in}																	
\setlength{\cellWidthb}{\textwidth/3-2\tabcolsep+0.40in}																	
\setlength{\cellWidthc}{\textwidth/3-2\tabcolsep+0.4in}																	
\scalebox{1}[1]{\begin{tabularx}{\textwidth}{>{\raggedright\arraybackslash}m{\cellWidtha}>{\raggedright\arraybackslash}m{\cellWidthb}>{\raggedright\arraybackslash}m{\cellWidthc}}																	

\toprule
\textbf{Househols}	& \textbf{Lockdown} & \textbf{Non-Lockdown}\\
\midrule 
EV1	&2018/07/22 00:00 to 2020/07/21 11:00	&2018/07/22 00:00 to 2020/02/29 23:00\\
EV2	&2018/07/22 00:00 to 2020/07/21 00:00	&\\
EV3	&2018/07/22 00:00 to 2020/07/21 10:00	&\\
EV4	&2018/07/22 00:00 to 2020/07/21 09:00	&\\

\bottomrule 
\end{tabularx}}

\end{table}

%
%
%

All experiments were coded in Python with the use of the PyTorch machine learning framework and the Ray Tune library for model training and tuning. This remainder of this section consists of three subsections: first, the results of \mbox{Mann-Whitney} analysis are discussed, next the hyperparameter search space is defined and training behavior is summarized for model optimization, and finally, the predictive performance is analyzed.

\subsection {Statistical Test Results}

Two trials are completed for each household, one for the full dataset with lockdown data and the other for the subset without lockdown data. After initial preparations are completed according to the described methodology, the dataset for non-lockdown  was split into train and test sets, similar in proportions to those used for the complete dataset. The shift in behavior due to lockdowns was analyzed to determine if there was a statistically significant difference in the distribution of the electricity consumption between train and test sets for lockdown and non-lockdown conditions.

In order to interpret the results of the \mbox{Mann-Whitney}, the null hypothesis was established. The null hypothesis in this scenario is that there is no statistically significant difference between the training and test set for any of the datasets. For the significance level of 5\%, the null hypothesis was rejected for cases where the P-value of the \mbox{Mann-Whitney} is less than 0.05 (5.00 $\times~10^{-2}$).  

The \emph{p}-value results of the \mbox{Mann-Whitney} analysis comparing training and test datasets shown in Table \ref{tab:Mann-WhitneyPvalues} confirm that the null hypothesis can be rejected for all datasets, as they fall significantly below the threshold value of  5.00 $\times~10^{-2}$. Therefore, the \mbox{Mann-Whitney} results indicate that there is a statistically significant difference between the training and test datasets regardless of lockdowns for all households.

\begin{table}[H] 
\caption{Mann-Whitney \emph{p}-value results for the lockdown and non-lockdown datasets.\label{tab:Mann-WhitneyPvalues}}
\setlength{\cellWidtha}{\textwidth/3-2\tabcolsep-0.4in}																	
\setlength{\cellWidthb}{\textwidth/3-2\tabcolsep+0.20in}																	
\setlength{\cellWidthc}{\textwidth/3-2\tabcolsep+0.2in}	
\scalebox{1}[1]{\begin{tabularx}{\textwidth}{>{\raggedright\arraybackslash}m{\cellWidtha}>{\raggedright\arraybackslash}m{\cellWidthb}>{\raggedright\arraybackslash}m{\cellWidthc}}		
\toprule
\textbf{Househols}	& \textbf{\emph{p}-Values Lockdown} & \textbf{\emph{p}-Values Non-Lockdown}\\
\midrule
EV1	&1.66 $\times~10^{-23}$	&9.48 $\times~10^{-5}$ \\
EV2	&8.25 $\times~10^{-3}$	&9.33 $\times~10^{-19}$ \\
EV3	&5.64 $\times~10^{-188}$	&1.07 $\times~10^{-31}$ \\
EV4	&9.22 $\times~10^{-43}$	&3.40 $\times~10^{-8}$ \\
\bottomrule
\end{tabularx}}
\end{table}

\subsection{Model Training and Tuning}

For each of the datasets outlined in Table \ref{tab:datasetDates}, 80 models were considered using Bayesian optimization within a defined hyperparameter search space. The hyperparameters tuned for the model were batch size, window size, the number of hidden layers, the number of neurons in the hidden layers, the Adam optimizer learning rate, and the dropout probability. The defined search space for each of the hyperparameters is summarized in Table \ref{tab:hyperparameters}. Every model that was trained had its performance evaluated using the performance metrics outlined in Section \ref{sec:PerformanceMetrics}.

\begin{table}[H] 
\caption{Hyperparameter search space.\label{tab:hyperparameters}}
\setlength{\cellWidtha}{\textwidth/2-2\tabcolsep-0.5in}																	
\setlength{\cellWidthb}{\textwidth/2-2\tabcolsep+0.5in}	 																
\scalebox{1}[1]{\begin{tabularx}{\textwidth}{>{\raggedright\arraybackslash}m{\cellWidtha}>{\raggedright\arraybackslash}m{\cellWidthb}}		
\toprule
\textbf{Hyperparameter}	& \textbf{Search Space} \\
\midrule
Batch Size	&32, 64, or 128 \\
Window Size	&12, 24, 48, or 72 \\
Hidden Layers	&1, 2, or 3 \\
Hidden Neurons	&64 or 128  \\
Learning Rate	&1 $\times~10^{-4}$, 1 $\times~10^{-3}$, or 1 $\times~10^{-2}$  \\
Dropout Probability	&5 $\times~10^{-2}$, 1 $\times~10^{-1}$, or 1.5 $\times~10^{-1}$ \\
\bottomrule
\end{tabularx}}
\end{table}

%

The input and output layers were each set to a fixed size. The number of neurons in the input layer was set by the number of features in the input dataset, and the output layer has a single neuron for the regression prediction output. Different sliding window sizes $w$ were used to provide the model with varying numbers of previous time steps to use as inputs. The window size is an important consideration because using a different number of previous time steps may help the models capture distinct patterns in each household’s electricity consumption. Each model predicts the energy consumption one hour ahead. 

\begin{figure}[H]
\includegraphics[width=8.5 cm]{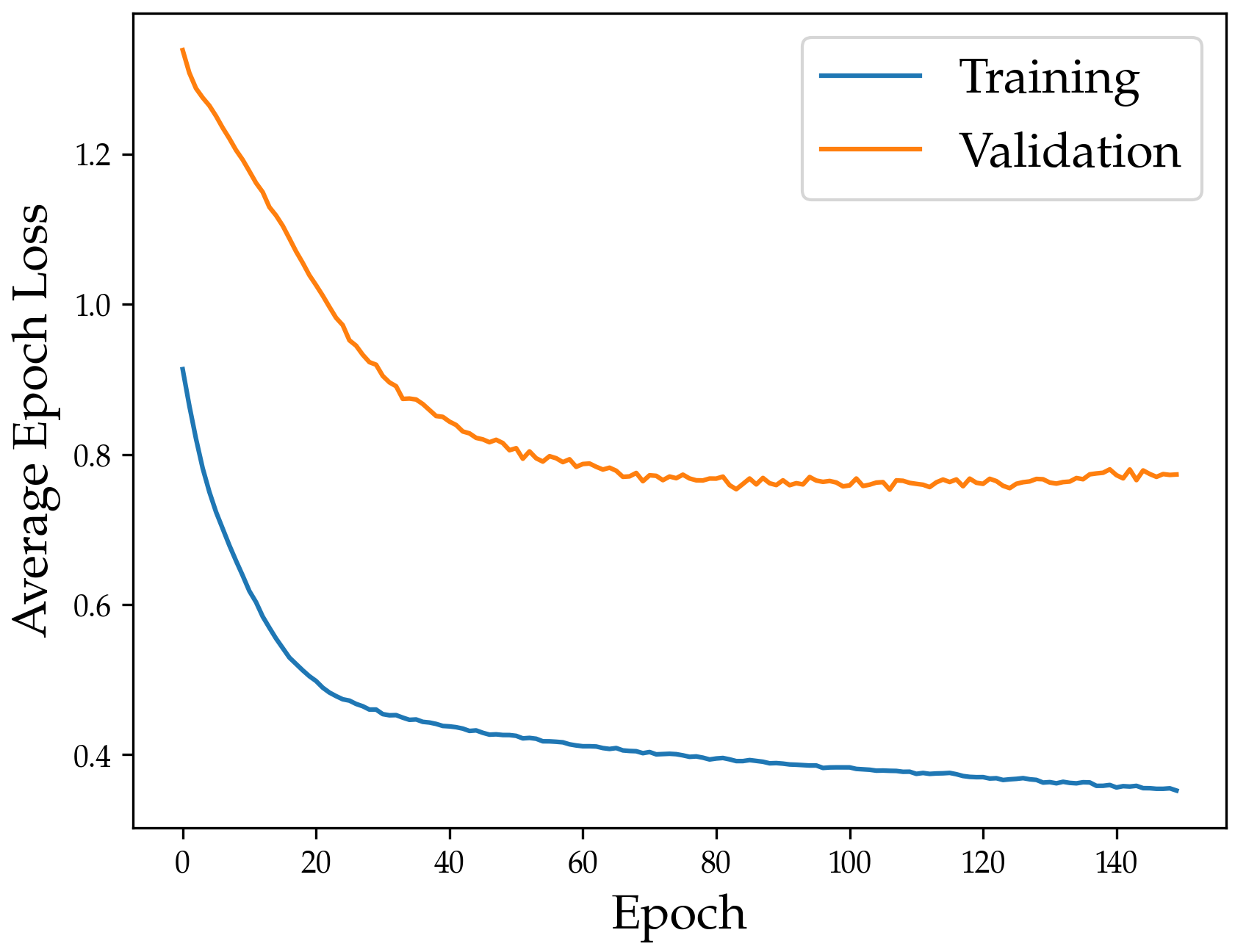}
\caption{Training and validation loss of the optimal LSTM-BNN EV3 non-lockdown model.\label{fig:loss}}
\end{figure}

The options for dropout probability tuning were based on the tuning range used by Zhang and Mahadevan \cite{10zhang2020bayesian} for creating BNN models. A dropout of zero was also included to act as a benchmark for how a non-BNN neural network would perform for each of the households. The hidden layer space and an epoch of 150 were determined by referring to the hyperparameters used for LSTM models for electricity load forecasting used by Kong et al. \cite{9kong2017short}. A sample of the training behavior of the optimal LSTM-BNN for EV3 and the non-lockdown dataset model is shown in Figure \ref{fig:loss}. From this figure, it can be seen that the early stopping could be beneficial in the reduction of computational resources, as a very minor improvement of the validation performance can be observed beyond approximately 80 epochs.

In experiments, the proposed LSTM-BNN interval forecasting is compared to the point forecasting. Both use LSTM as their base model and both undergo exactly the same dataset preparation, preprocessing, and training and tuning steps, as described in Sections \mbox{\ref{sec:DatasetPreparation}} and \mbox{\ref{sec:preprocessing}}, and \mbox{\ref{sec:TrainingTuning}}, respectively. The difference is that in point forecasting, at inference time, the dropout is not active, and, therefore, point forecasting results in a single precision for each time step. In contrast, the proposed LSTM-BNN generates multiple predictions and forms an interval with the BNN technique. 

\subsection{Predictive Performance Analysis}

The performance results are explored in three parts: first, the overall performance of point and interval prediction models are examined, followed by the analysis of the performance among households. Next, interval forecasts are compared to point forecasts, and lockdown is compared to non-lockdown. Finally, the correlation between Mann-Whitney results and model performance is examined.

\subsubsection{Overall Performance}
Table \ref{tab:modelComparison} shows the average MAPE values for the four households for each of the two datasets, lockdown and non-lockdown, and for each of the two approaches, point and interval prediction approaches. For interval forecasts, the four performance metrics were calculated using the mean of the generated interval from the set of the forecast generated with the Bayesian technique as described in Section \ref{sec:IntervalForecasting}.     
For each household, dataset, and point/interval approach, the model was tuned, and the results from the tuned models were averaged and reported in this table. All MAPE values are significantly higher than those reported in the literature \cite{8fekri2022distributed, 12fekri2021deep}, but that is to be expected as EV charging behavior adds remarkable variability and randomness to power consumption pattern compared with office buildings or households without EVs. In general, excluding electricity consumption lockdown data from March 2020 or later did not create an improvement in the predictive performance of the models. While an increase in the error between actual and predicted values is expected between training and testing, there is a much greater increase for the non-lockdown dataset than for the lockdown dataset.

The average interval prediction performance shows that model performance was better overall on the full dataset that included lockdown data. Point predictions for the lockdown were also better than for the non-lockdown for the test dataset. This result is somewhat surprising considering that it would be expected that electricity consumption would be more difficult to predict in a lockdown environment than in a non-lockdown environment when the historical data used for creating models are based mainly on non-lockdown behavior. A possible reason for this is that with the full dataset, the model has more data to learn from.  

\begin{table}[H] 
 
\caption{Average model performance~comparison.\label{tab:modelComparison}}
\setlength{\cellWidtha}{\textwidth/5-2\tabcolsep-0.0in}											
\setlength{\cellWidthb}{\textwidth/5-2\tabcolsep-0.1in}											
\setlength{\cellWidthc}{\textwidth/5-2\tabcolsep-0in}											
\setlength{\cellWidthd}{\textwidth/5-2\tabcolsep+0.2in}											
\setlength{\cellWidthe}{\textwidth/5-2\tabcolsep-0.1in}											
\scalebox{1}[1]{\begin{tabularx}{\textwidth}{>{\raggedright\arraybackslash}m{\cellWidtha}>{\raggedright\arraybackslash}m{\cellWidthb}>{\raggedright\arraybackslash}m{\cellWidthc}>{\raggedright\arraybackslash}m{\cellWidthd}>{\raggedright\arraybackslash}m{\cellWidthe}}											

\toprule
\textbf{Dataset}	& \textbf{Model Type} & \textbf{Train MAPE} & \textbf{Validation MAPE} & \textbf{Test MAPE}\\
\midrule
Lockdown	&Point	&44.8121	&62.2463	&63.4173\\
&	Interval	&37.0744	&59.8753	&68.7626 \\
\midrule
Non-Lockdown	&Point	&37.2093	&61.1446	&71.5252 \\
&	Interval	&44.1562	&68.2564	&122.3360 \\

\bottomrule
\end{tabularx}}
\end{table}

%
%

\begin{figure}[H]
\includegraphics[width=12.5 cm]{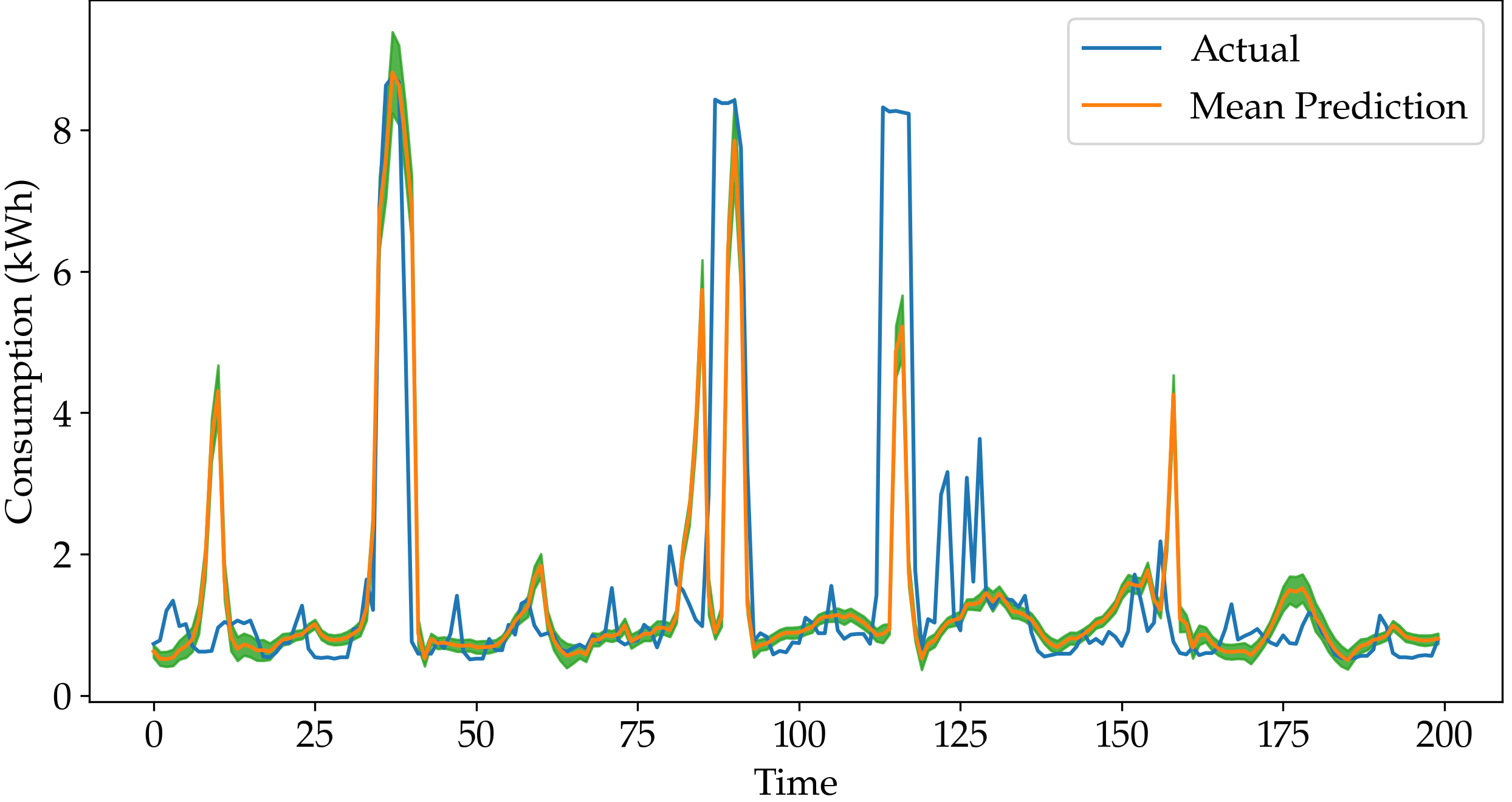}
\caption{The first 200 predictions made on the validation set of the optimized EV3 model for the lockdown dataset. The green areas indicate one standard deviation above and one standard deviation below the mean prediction value.\label{fig:predictions}}
\end{figure}

Figure \ref{fig:predictions} shows an example of interval forecast results; specifically actual energy consumption and predicted interval forecasts for EV3 and the lockdown dataset. It can be observed that the prediction interval varies throughout time indicating uncertainties in the forecasted values.  

While the focus of this work is the comparison of interval and point predictions, here we further examine interval forecast for the example EV3. Prediction Interval Coverage Probability (\mbox{$PICP$}) measures the fraction of actual values that lay within the prediction interval. In the case of the proposed LSTM-BNN, this will vary depending on how many standard deviations \mbox{$\sigma$} are forming the interval. For the interval of \mbox{$\pm 1\sigma$, $PICP$} is only 0.28, while increasing to \mbox{$\pm 2\sigma$, $\pm 3\sigma$, and $\pm 5\sigma$, increases $PICP$} to 0.51, 0.68, 0.82, respectively. Note that similar to MAPE values from Table \mbox{\ref{tab:modelComparison}}, these \mbox{$PICP$} values indicate high errors caused by the randomness of EV charging.

\subsubsection{Performance Among Households}

The results for the best-performing model for each individual household, for point and interval forecasts, are shown in Tables \ref{tab:lockdown} and \ref{tab:non-lockdown} for lockdown and non-lockdown datasets respectively. Note that models are created for each household individually, and not for a group of homes. The results include all four metrics MAPE, MSE, RMSE, and MAE for all datasets train, validation, and test. The best-performing point and interval prediction models in terms of test set MAPE were for households EV3 and EV4 respectively on the lockdown dataset. Although there are some cases where the models generate better predictions, such as test MAPE of about 31\% for EV3 with lockdown dataset and point predictions, most others exhibit higher error. Despite the success of LSTM models in forecasting electricity consumption for offices and households \mbox{\cite{8fekri2022distributed,218781349}}, the results show that in the presence of EV charging their accuracy is greatly reduced. As noted by other studies \mbox{\cite{8fekri2022distributed,9kong2017short}} the variability in household electricity consumption makes creating accurate predictions at this granularity challenging, which seems to only be exacerbated further with the additional consideration of EV charging. This variability results in much higher MAPE, irrelevant of the dataset or the household, than those typically reported in the literature. However, the literature commonly considers offices, schools, or a group of buildings that have much more predictable energy consumption patterns. Moreover, MAPE can produce very high values when the actual values are close to zero. Note that, while MSE, RMSE, and MAE are included in Tables \ref{tab:lockdown} and \ref{tab:non-lockdown}, we are not comparing among households based on those metrics as they are dependent on the scale of the actual values. 

\begin{table}[H] 
\caption{Lockdown~results.\label{tab:lockdown}}
\resizebox{\textwidth}{!}{	 
\begin{tabular}{llrrrrrrrr}
\toprule
&&\multicolumn{8}{c}{\textbf{Household}}\\
\cmidrule{3-10}  \\ [-6pt]
&  & \multicolumn{2}{c}{\textbf{EV1}} & \multicolumn{2}{c}{\textbf{EV2}} & \multicolumn{2}{c}{\textbf{EV3}} & \multicolumn{2}{c}{\textbf{EV4}}\\ 
\cmidrule{3-10} \\ [-6pt] 
\multirow{-3}{*}{\vspace{22pt}\textbf{Metric}	}&\multirow{-3}{*}{\vspace{22pt}\textbf{Model Type}	}&\textbf{Point}	&\textbf{Interval}	&\textbf{Point}	&\textbf{Interval}	&\textbf{Point}	&\textbf{Interval}	&\textbf{Point}	&\textbf{Interval}\\
\midrule
MAPE	&Train 	&70.0221	&35.7423	&43.0054	&56.4060	&26.7295	&25.0654	&39.4916	&31.0839\\
	&Validation	&78.0701	&80.1722	&92.6025	&87.0372	&36.3606	&37.5774	&41.9519	&34.7142 \\
	&Test	&47.0062	&70.9079	&122.2161	&103.2104	&31.5572	&57.0495	&52.8898 	&43.8828\\ 
\midrule
MSE	&Train 	&0.3033	&0.0470	&0.2293	&0.3027	&0.3726	&0.3395	&0.7405	&0.3801\\
	&Validation	&0.2600	&0.2528	&0.8229	&0.7407	&1.3065	&1.4637	&1.1385	&1.2946\\
	&Test	&0.4742	&0.4492	&2.0359	&1.8019	&1.5855	&4.2337	&1.8502	&1.9896\\
\midrule
RMSE	&Train 	&0.5507	&0.2168	&0.4789	&0.5502	&0.6104	&0.5827	&0.8605	&0.6165\\
	&Validation	&0.5099	&0.5028	&0.9071	&0.8606	&1.1430	&1.2098	&1.0670	&1.1378\\
	&Test	&0.6886	&0.6703	&1.4269	&1.3423	&1.2592	&2.0576	&1.3602	&1.4105\\
\midrule
MAE	&Train 	&0.2687	&0.1190	&0.2870	&0.3348	&0.3281	&0.3152	&0.4771	&0.3519\\
	&Validation	&0.2089	&0.2121	&0.5672	&0.5249	&0.4907	&0.5043	&0.5545	&0.4847\\
	&Test	&0.3617	&0.3758	&0.7978	&0.7444	&0.6656	&1.1306	&0.7871	&0.6899\\

\bottomrule
\end{tabular}}
\end{table}
\unskip

\begin{table}[H] 
\caption{Non-lockdown~results.\label{tab:non-lockdown}}
\resizebox{\textwidth}{!}{	
\begin{tabular}{llrrrrrrrr}
\toprule
&&\multicolumn{8}{c}{\textbf{Household}}\\
\cmidrule{3-10}  \\ [-6pt]
&  & \multicolumn{2}{c}{\textbf{EV1}} & \multicolumn{2}{c}{\textbf{EV2}} & \multicolumn{2}{c}{\textbf{EV3}} & \multicolumn{2}{c}{\textbf{EV4}}\\ 
\cmidrule{3-10} \\ [-6pt]
\multirow{-3}{*}{\vspace{22pt}\textbf{Metric}}	&\multirow{-3}{*}{\vspace{22pt}\textbf{Model Type}}	&\textbf{Point}	&\textbf{Interval}	&\textbf{Point}	&\textbf{Interval}	&\textbf{Point}	&\textbf{Interval}	&\textbf{Point}	&\textbf{Interval}\\
\midrule
MAPE	&Train 	&71.9591	&71.3261	&26.4764	&35.0559	&32.2495	&33.4092	&18.1520	&36.8335 \\
	&Validation	&79.8842	&116.1025	&79.5442	&81.3031	&39.4762	&35.7723	&45.6739	&39.8479\\
	&Test	&84.0878	&266.6882	&62.2114	&72.7572	&65.6180	&94.3859	&74.1836	&55.5127\\
\midrule
MSE	&Train 	&0.9714	&0.4387	&0.0768	&0.1281	&0.4747	&0.8219	&0.0845	&0.8166\\
	&Validation	&0.3199	&0.4667	&1.1516	&1.2964	&2.3181	&1.8993	&2.0753	&1.7791\\
	&Test	&0.5117	&0.5249	&0.8616	&0.8402	&2.7516	&3.1830	&2.3104	&1.2918\\
\midrule
RMSE	&Train 	&0.9856	&0.6623	&0.2771	&0.3580	&0.6890	&0.9066	&0.2908	&0.9037\\
	&Validation	&0.5656	&0.6832	&1.0731	&1.1386	&1.5225	&1.3781	&1.4406	&1.3338\\
	&Test	&0.7153	&0.7245	&0.9282	&0.9166	&1.6588	&1.7841	&1.5200	&1.1366\\
\midrule
MAE	&Train 	&0.4498	&0.3085	&0.1641	&0.2026	&0.3912	&0.4663	&0.1792	&0.4672\\
	&Validation	&0.2312	&0.3241	&0.6561	&0.6889	&0.6265	&0.5881	&0.7280	&0.6617\\
	&Test	&0.2681	&0.4952	&0.5562	&0.5458	&0.7869	&0.9827	&0.8580	&0.6530\\

\bottomrule
\end{tabular}}
\end{table}

\begin{figure}[H]
     {
 \captionsetup{justification=centering}
    \begin{subfigure}[b]{0.48\textwidth}
         \centering
         \includegraphics[width=\textwidth]{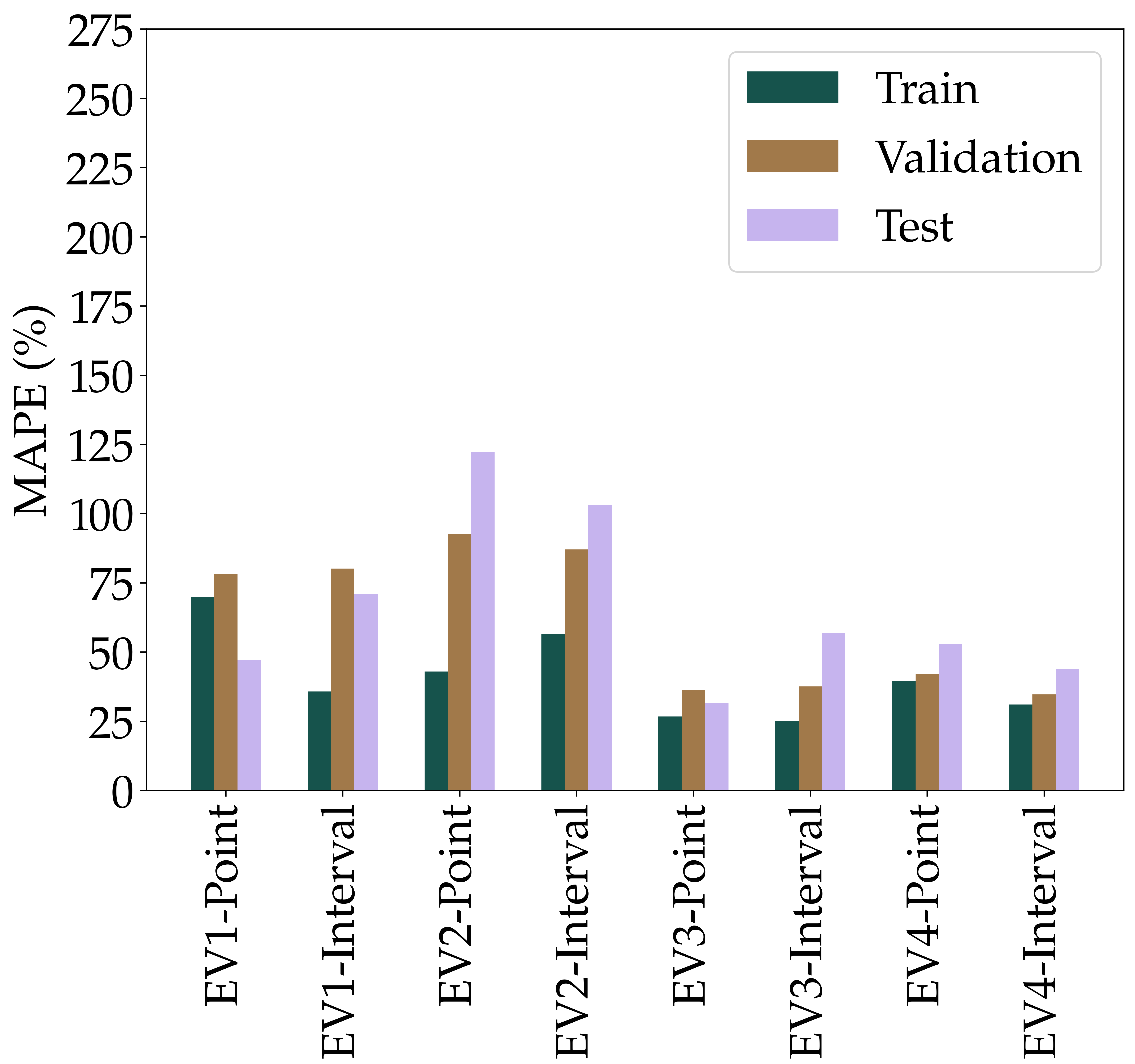}
          \caption{MAPE values for lockdown~data.} \label{fig:MAPE-lockdown}
     \end{subfigure}
     \hfill
   \captionsetup{justification=centering}
  \begin{subfigure}[b]{0.48\textwidth}
         \centering
          \includegraphics[width=\textwidth]{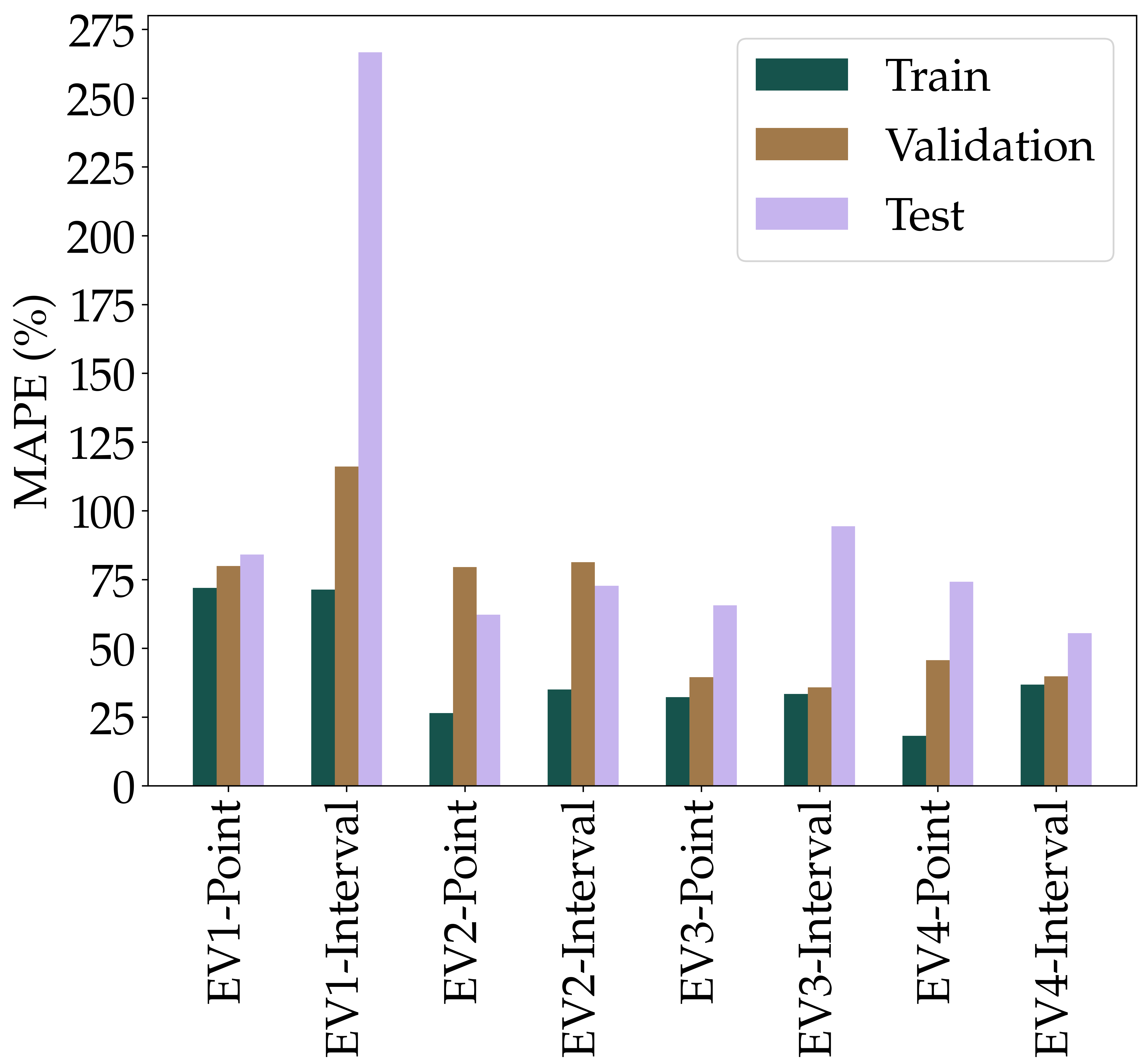}
         \caption{MAPE values for non-lockdown~data} \label{fig:MAPE-nonlockdown}
         \end{subfigure}}
    \caption{MAPE values for lockdown and non-lockdown~data.\label{fig:MAPE}}
\end{figure}

To analyze household differences, Figure \ref{fig:MAPE-lockdown} shows the MAPE values for point and interval predictions for each household for lockdown and lockdown data while Figure \ref{fig:MAPE-nonlockdown} does the same for the non-lockdown scenario. It can be observed that there are some households that were easier for the predictive models to capture. The EV3 and EV4 household datasets produced the best-performing models but the models trained on the EV1 and EV2 households were much less successful. The predictive performance of the model trained on the EV2 lockdown dataset which produces the highest \emph{p}-value highlights that there is no direct relationship between \mbox{Mann-Whitney} results and model performance. The Mann-Whitney results from Table \ref{tab:Mann-WhitneyPvalues} indicate that in all scenarios, there is a significant difference between the training and test dataset which could be one of the reasons for the high MAPE results. 

%

\begin{figure}[H]
 
 \includegraphics[width=11 cm]{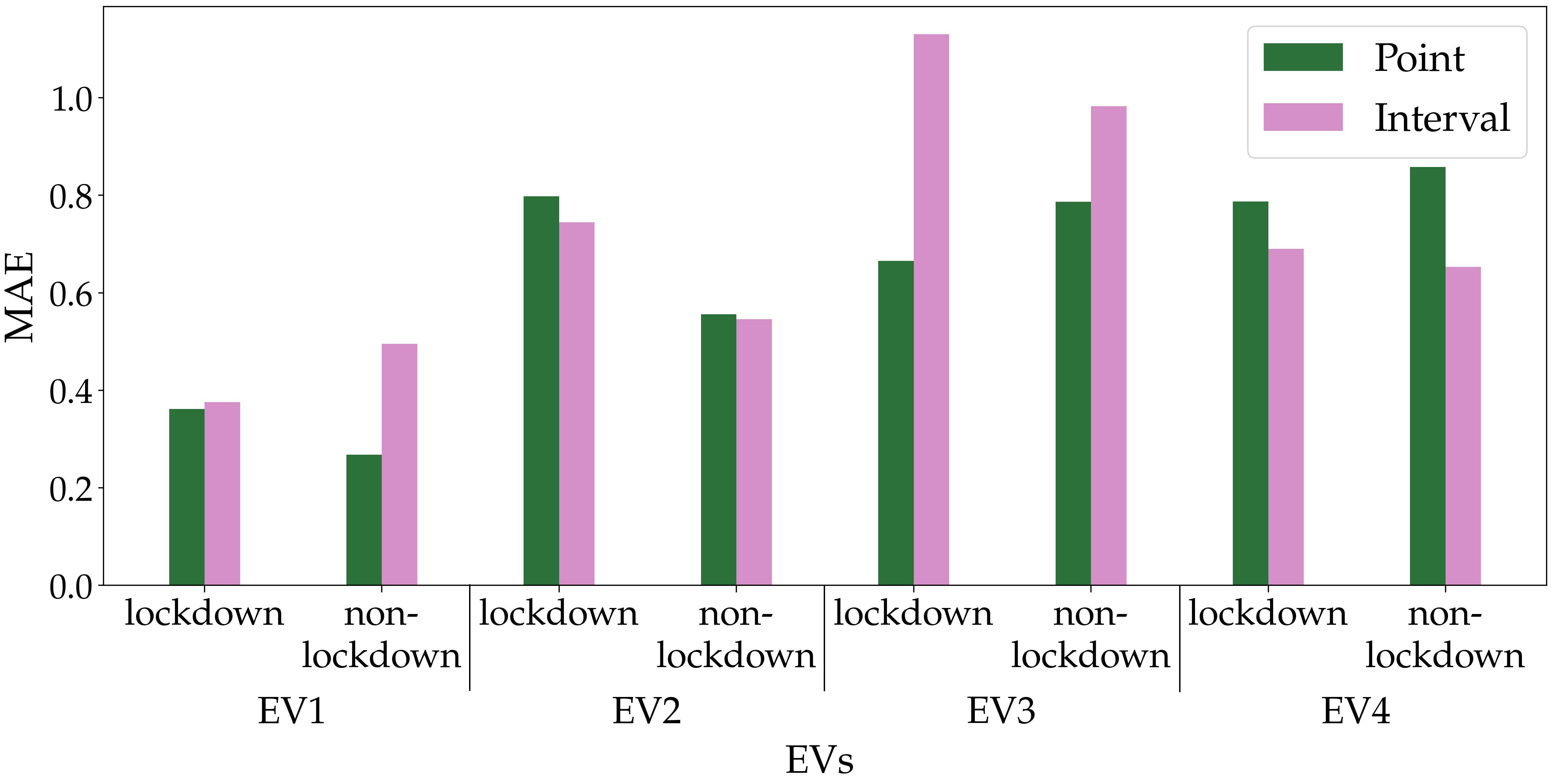}
\caption{{MAE values for lockdown and non-lockdown~data.}\label{fig:MAE}}
\end{figure}

To analyze this data from a different perspective, Figure \mbox{\ref{fig:MAE}} shows the MAE values for the four EVs for point and interval prediction in lockdown and non-lockdown periods. Since MAE is scale dependent and consumption scales vary among households, different approaches should be compared for each household individually and not among households. For four scenarios, point forecasting achieves lower error than interval forecasting; however, interval forecasting has the advantage of providing uncertainty information. 

\subsubsection{Interval Forecasts Compared to Point Forecasts}
\begin{figure}[H]
     \centering
     \begin{subfigure}[b]{0.48\textwidth}
         \centering
         \includegraphics[width=\textwidth]{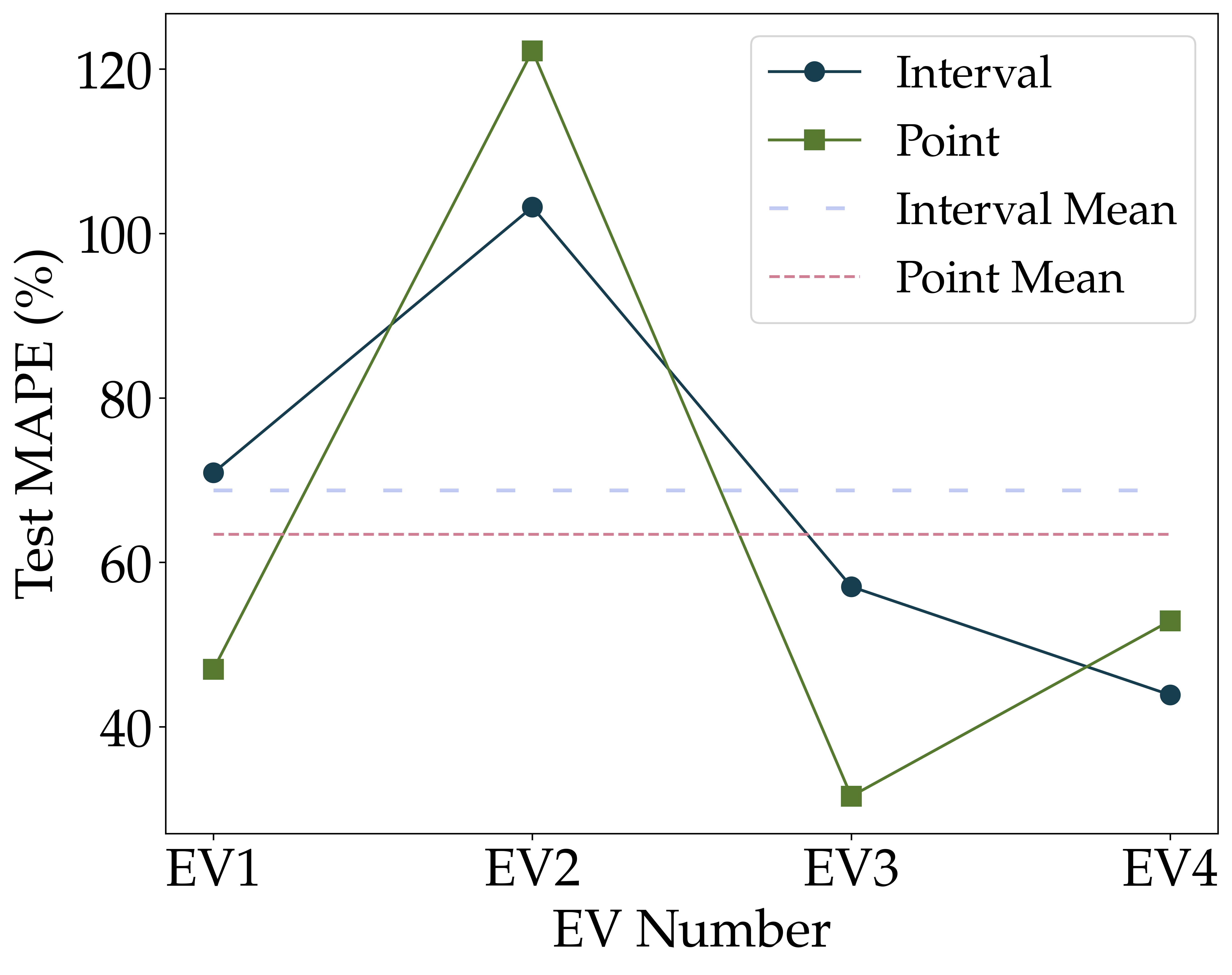}
          \caption{Comparison for lockdown} \label{fig:MAPE-P-R-lockdown}
     \end{subfigure}
     \hfill
     \begin{subfigure}[b]{0.48\textwidth}
         \centering
         \includegraphics[width=\textwidth]{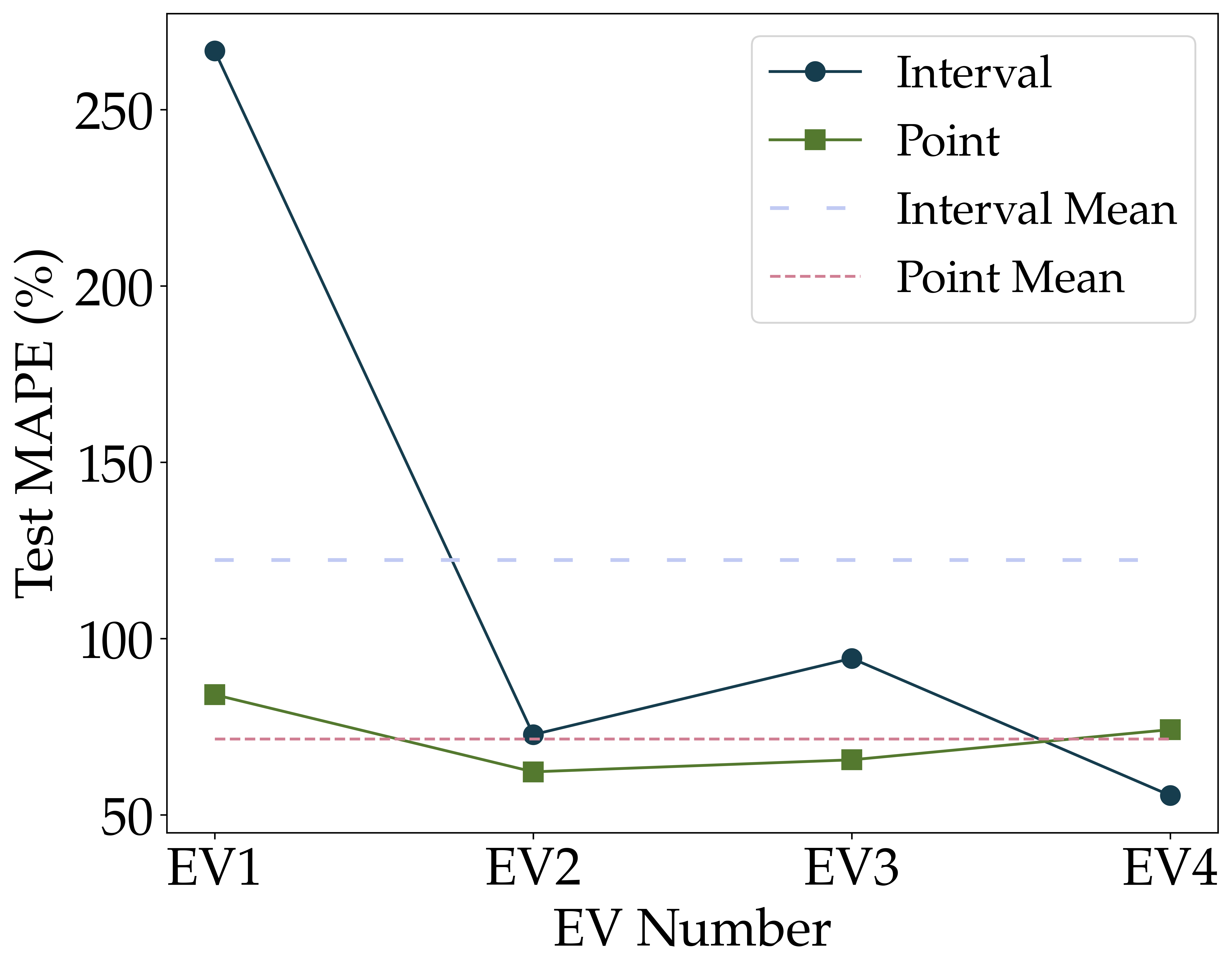}
         \caption{Comparison for non-lockdown} \label{fig:MAPE-P-N-nonlockdown}
         \end{subfigure}
    
    \caption{Comparison between interval and point forecasts for lockdown and non-lockdown data.\label{fig:MAPE-P-R-value}}
\end{figure}

Figure \ref{fig:MAPE-P-R-lockdown} compares interval forecasts to point forecasts in terms of MAPE for the lockdown test dataset. In addition to point and interval MAPE for each household, it includes the averages for the four households. It can be observed that the average MAPE for the point forecast is about 5\% lower than the average MAPE for interval forecasts. Also, this figure highlights that EV2 was the hardest household to predict for both point and interval approaches. At the same time, EV3 was the most straightforward prediction for point forecasting while EV4 achieved the lowest prediction error for interval forecasting. Figure \ref{fig:MAPE-P-N-nonlockdown} shows the same comparison but for non-lockdown test data. Again, MAPE for interval forecasting is higher than for point forecasting but the difference between average MAPE for interval and point forecasts is much larger for the non-lockdown dataset than for the lockdown dataset. Interval forecasting for EV1 performed especially poorly which raised the average MAPE for interval forecasts. 

\subsubsection{Lockdown Compared to Non-lockdown}

\begin{figure}[H]
     \begin{subfigure}[b]{0.45\textwidth}
         \centering
         \includegraphics[width=\textwidth]{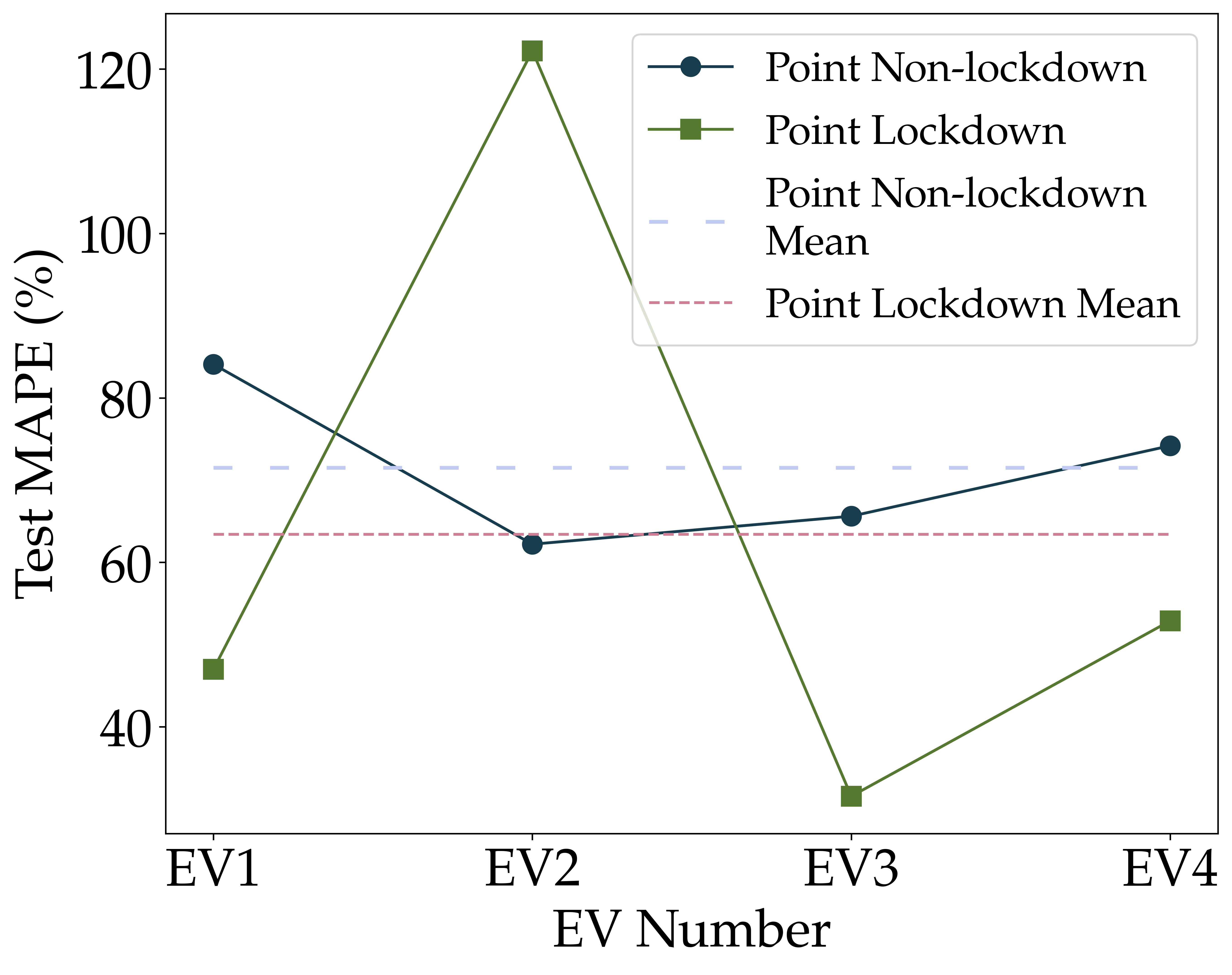}
          \caption{Comparison for point forecasting} \label{fig:MAPE-P-N-lockdown}
     \end{subfigure}
     \hfill
     \begin{subfigure}[b]{0.45\textwidth}
         \centering
         \includegraphics[width=\textwidth]{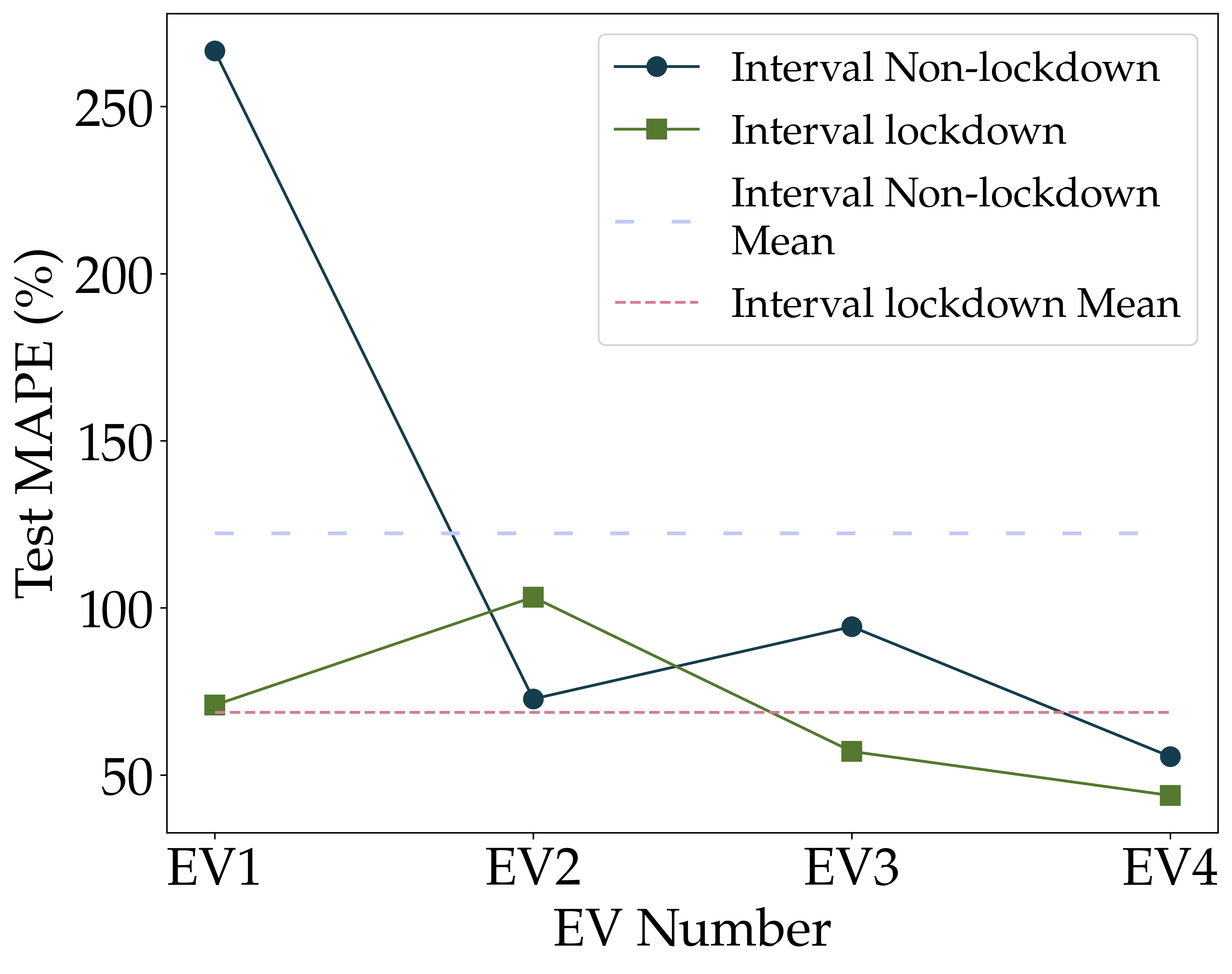}
         \caption{Comparison for interval forecasting} \label{fig:MAPE-R-N-lockdown}
    \end{subfigure}
    
    \caption{Comparison between lockdown and non-lockdown results for point and interval forecasts.\label{fig:MAPE-P-value}}
\end{figure}

Figure \ref{fig:MAPE-P-N-lockdown} and Figure \ref{fig:MAPE-R-N-lockdown} compare forecasting with lockdown and non-lockdown for point and interval forecasting  respectively. Figure \ref{fig:MAPE-P-N-lockdown} shows that average point MAPE for non-lockdown data is about 8\% higher than for lockdown. All lockdown MAPE values except EV2 were much lower than non-lockdown predictions. Similarly, Figure \ref{fig:MAPE-R-N-lockdown} indicates that interval prediction for lockdown achieves better results than for non-lockdown except EV2. Overall, accuracy varies greatly among households.

\subsubsection{Analysis of Relation between Predictive Performance and Mann-Whitney Results}

\begin{figure}[H]
 
 {   \captionsetup{justification=centering}
 \begin{subfigure}[b]{0.48\textwidth}
         \centering
          \includegraphics[width=\textwidth]{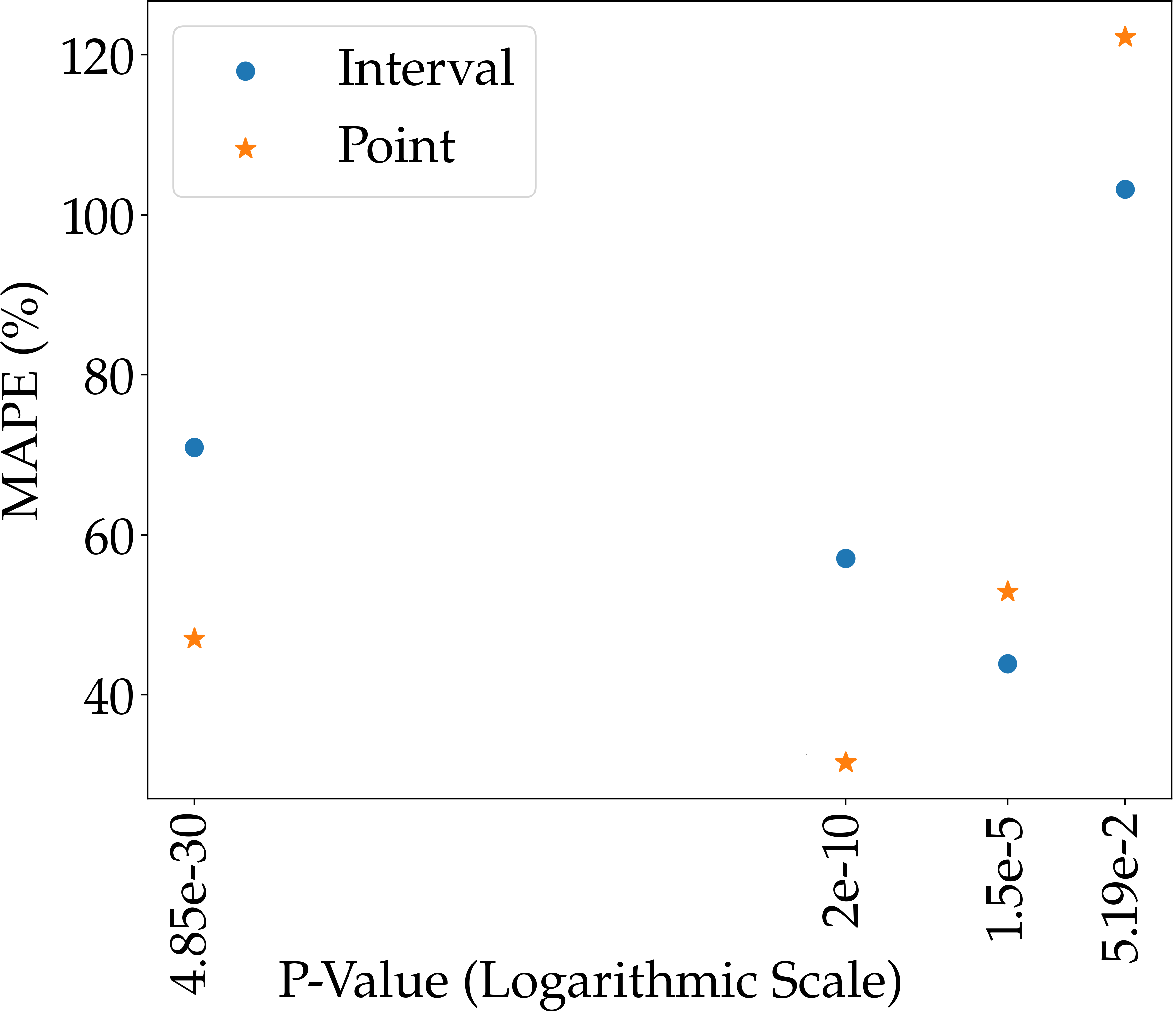}
          \caption{~} \label{fig:MAPE-P-lockdown}
     \end{subfigure}
     \hfill
     \begin{subfigure}[b]{0.48\textwidth}
         \centering
         \includegraphics[width=\textwidth]{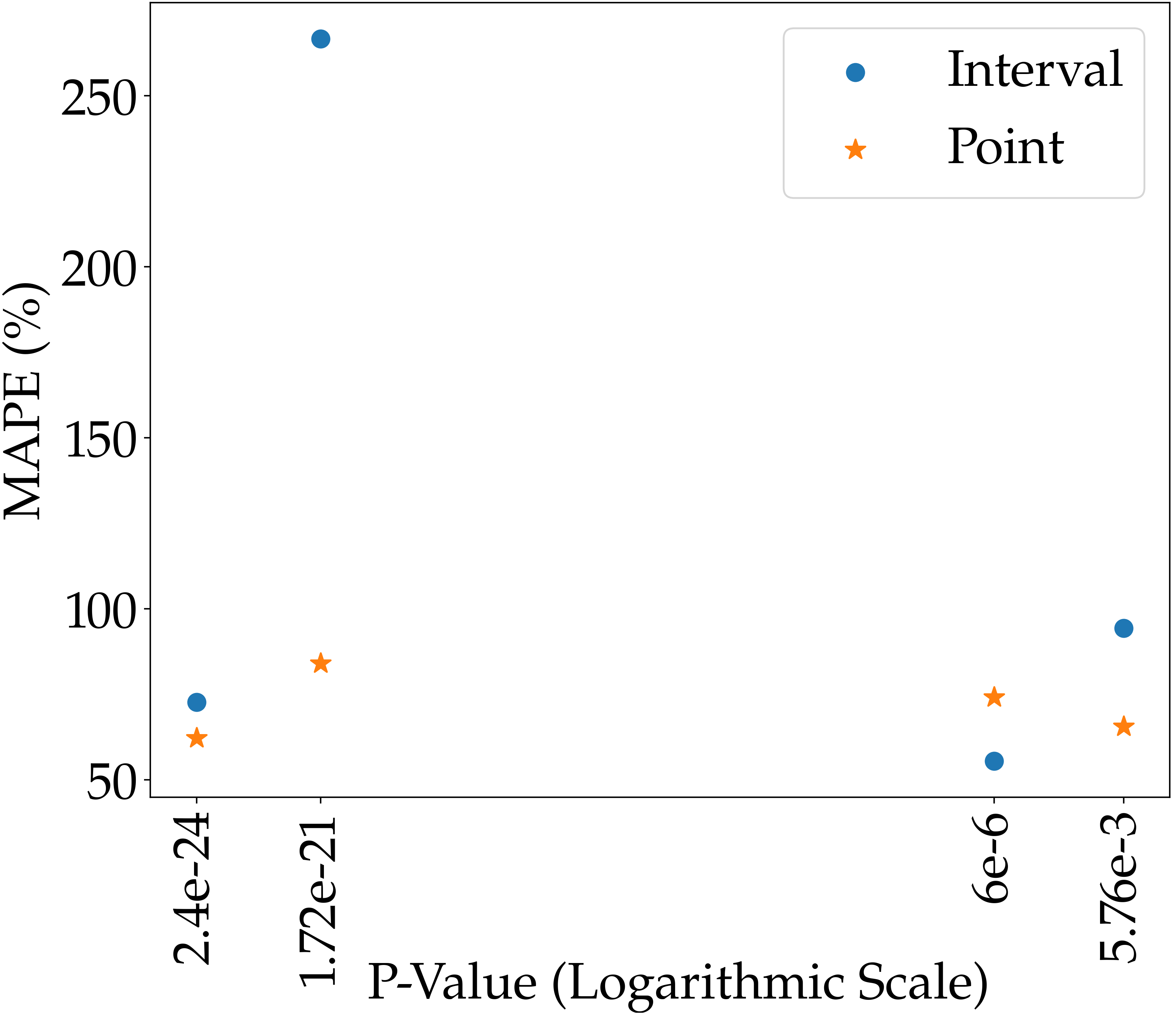}
      \captionsetup{justification=centering}
\caption{~} \label{fig:MAPE-P-nonlockdown}
         \end{subfigure}   }
    \caption{MAPE performance compared to Mann-Whitney P-Value significance~measure. (\textbf{a}) Lockdown MAPE performance compared to Mann-Whitney \emph{p}-Value significance~measure;   (\textbf{b}) Non-lockdown MAPE performance compared to Mann-Whitney \emph{p}-Value significance~measure. \label{fig8}}
\end{figure}

Figure \ref{fig:MAPE-P-lockdown} and Figure \ref{fig:MAPE-P-nonlockdown} relate model performance on the test set (MAPE) and \linebreak \mbox{Mann-Whitney} P-value between training and test data. These figures further demonstrate that there is no conclusive relationship between the \mbox{Mann-Whitney} P-value and model performance on the test set. This is due to \mbox{Mann-Whitney} comparing the datasets statistically while the prediction performance depends on many other factors such as the quality of features and randomness of data. Still, the Mann-Whitney test demonstrated that there is a significant difference between the train and test datasets, regardless of whether lockdown data were included or not, which is a contributing factor to the higher error values observed. 

\unskip


\section{Conclusions}

Accurately predicting electricity consumption is an important factor in providing an adequate and reliable energy supply. The electricity demand will continue to grow as society transitions away from using ICE vehicles to EVs that are able to be charged in residential homes. However, predicting energy consumption at the individual household level is more challenging than forecasting for office buildings, schools, or regions due to the high variability in electricity consumption patterns\cite{13zhang2018forecasting, grolinger2016energy}. This challenge is further amplified by the need to accommodate EV charging. 

This paper proposes LSTM-BNN interval load forecasting for individual households in presence of EV charging based on the LSTM deep learning model and Bayesian inference. The LSTM model incorporates the dropout layer which is active during the inference time and responsible for generating a set of point predictions for a single input sample. Then, the Bayesian technique is employed to create interval forecasts from this set of predictions. The achieved accuracy varies greatly among households due to the variability and randomness of their energy consumption patterns. Examining the performance of point and interval prediction models shows that the LSTM-BNN interval prediction model performs similarly to a standard LSTM point prediction model with the benefit of providing an interval for the prediction. Although the proposed LSTM-BNN is more complex and involves longer training time than the traditional LSTM point forecasting models, LSTM-BNN predictions quantify uncertainly and offer additional information for decision-making. This paper also examined the impact of the COVID-19 lockdown on the load forecasting for these households: results show that the proposed LSTM-BNN achieves similar results for lockdown and non-lockdown periods. We stipulate that the randomness of the EV charging patterns outweighs the impact of change due to the lockdowns.

As demonstrated in our study, EV charging is highly variable and predicting household energy consumption in presence of EV charging is difficult. For use cases such as infrastructure planning, forecasting energy consumption for a neighborhood block may be sufficient. For such scenarios, aggregating energy consumption on the block level would remove some of the randomnesses and improve forecasting accuracy.  
Future work will examine the results in terms of the size of the prediction interval to be able to better relate different interval forecasts. Moreover, alternative methods to \mbox{Mann-Whitney} will be considered to acquire better insight into the potential changes in consumption habits during different periods of time. As energy consumption patterns, including EV charging patterns, change over time resulting in what is known as concept drift, the techniques such as online learning \mbox{\cite{12fekri2021deep}} could be integrated with the proposed approach to better capture changes over time.

\vspace{6pt} 



\authorcontributions{Conceptualization, R.S., S.M.; methodology, R.S., K.G.; software, R.S.; validation, R.S.; formal analysis, R.S., M.E., K.G.; investigation, R.S., M.E., K.G.; resources, K.G. and S.M.; writing—original draft preparation, R.S.; writing—review and editing, K.G., M.E., S.M.; visualization, R.S., M.E.; supervision, K.G.; project administration, K.G.; funding acquisition, K.G., S.M. All authors have read and agreed to the published version of the manuscript.}

\funding{This research has been supported by Ontario Centre of Innovation under grant OCI \#34674 and by the Natural Sciences and Engineering Research Council of Canada (NSERC) under grant ALLRP 570760-21. Computation was enabled in part by the Digital Research Alliance of Canada.}

\institutionalreview{Not applicable.}

\informedconsent{Not applicable.}

\dataavailability{Data analyzed in this study are obtained from London Hydro and are protected under a signed non-disclosure agreement. Approvals are needed for sharing this data.} 

\acknowledgments{The authors would like to thank London Hydro for supplying industry knowledge and data used in this study.}

\conflictsofinterest{The authors declare no conflict of interest.} 



\begin{adjustwidth}{-\extralength}{0cm}

\reftitle{References}

\PublishersNote{}
\end{adjustwidth}
\end{document}